\documentclass[pdflatex,sn-mathphys-num]{sn-jnl}
\usepackage{graphicx}%
\usepackage{microtype}%
\usepackage{array}
\usepackage{framed}
\usepackage{mdframed}
\usepackage{makecell}
\usepackage{multirow}%
\usepackage{amsmath,amssymb,amsfonts}%
\usepackage{amsthm}%
\usepackage{mathrsfs}%
\usepackage[noabbrev,nameinlink]{cleveref}
\usepackage[title]{appendix}%
\usepackage[dvipsnames]{xcolor}%
\usepackage{textcomp}%
\usepackage{fix-cm}  
\usepackage{manyfoot}%
\usepackage{booktabs}%
\usepackage{lineno}
\usepackage{algorithm}%
\usepackage[noEnd,indLines,commentColor=gray]{algpseudocodex}
\AddToHook{env/algorithmic/begin}{\tikzexternaldisable}
\usepackage{tikz}
\usetikzlibrary{external}
\usepackage{siunitx}
\usepackage{pgfplots}
\usepackage{pgfplotstable}
\usepackage{geometry}
\usepackage{csvsimple}
\usepackage{ifthen}
\usepackage[acronym]{glossaries}
\glsdisablehyper%
\usepackage[export]{adjustbox}
\usepackage{hyperref}
\usetikzlibrary{spy}
\usetikzlibrary{shapes.misc, positioning}
\usepgfplotslibrary{statistics}

\AtBeginEnvironment{appendices}{%
  \crefalias{section}{appendix}%
}

\AtEndEnvironment{appendices}{%
  \crefalias{section}{section}%
}

\pgfplotsset{compat=1.17}
\usepgfplotslibrary{%
	groupplots,
	external,
}

\pgfplotsset{
    boxplot/every average/.style={draw=orange, mark=diamond*, fill=orange, mark size=2pt},
    boxplot/every median/.style={draw=orange, fill=orange},
}

\tikzexternaldisable

\newcommand\norm[1]{\lVert#1\rVert}
\algrenewcommand{\algorithmicensure}{\textbf{Return}}

\pgfplotsset{colormap={flare}{%
rgb = (0.92539502, 0.64345456, 0.47594352)
rgb = (0.92077582, 0.59804722, 0.44818634)
rgb = (0.9155979, 0.55210684, 0.42070204)
rgb = (0.90921368, 0.5056543, 0.39544411)
rgb = (0.90077904, 0.45884905, 0.37556121)
rgb = (0.888292, 0.40830288, 0.36223756)
rgb = (0.87199254, 0.3633634, 0.35974223)
rgb = (0.84916723, 0.32289973, 0.36711424)
rgb = (0.81942908, 0.28911553, 0.38102921)
rgb = (0.7826624, 0.26420493, 0.39754146)
rgb = (0.73695678, 0.24620072, 0.41357737)
rgb = (0.69226314, 0.23413578, 0.42480327)
rgb = (0.64795375, 0.22217149, 0.43330852)
rgb = (0.60407977, 0.21017746, 0.43913439)
rgb = (0.56041794, 0.19845221, 0.44207535)
rgb = (0.51278481, 0.18693492, 0.44112605)
rgb = (0.46818879, 0.17788392, 0.43552047)
rgb = (0.42355299, 0.16934709, 0.42581586)
rgb = (0.37928736, 0.16052483, 0.41270599)
rgb = (0.33604378, 0.15006017, 0.39835754)
}}

\definecolor{flare0}{rgb}{0.92539502, 0.64345456, 0.47594352}
\definecolor{flare8}{rgb}{0.81942908, 0.28911553, 0.38102921}
\definecolor{flare17}{rgb}{0.42355299, 0.16934709, 0.42581586}

\pgfplotsset{colormap={inferno}{%
rgb = (1.46200e-03, 4.66000e-04, 1.38660e-02)
rgb = (2.94320e-02, 2.15030e-02, 1.14621e-01)
rgb = (9.29900e-02, 4.55830e-02, 2.34358e-01)
rgb = (1.83429e-01, 4.03290e-02, 3.54971e-01)
rgb = (2.71347e-01, 4.09220e-02, 4.11976e-01)
rgb = (3.60284e-01, 6.92470e-02, 4.31497e-01)
rgb = (4.41207e-01, 9.93380e-02, 4.31594e-01)
rgb = (5.28444e-01, 1.30341e-01, 4.18142e-01)
rgb = (6.09330e-01, 1.59474e-01, 3.93589e-01)
rgb = (6.94627e-01, 1.95021e-01, 3.54388e-01)
rgb = (7.69556e-01, 2.36077e-01, 3.07485e-01)
rgb = (8.41969e-01, 2.92933e-01, 2.48564e-01)
rgb = (8.98192e-01, 3.58911e-01, 1.88860e-01)
rgb = (9.44285e-01, 4.42772e-01, 1.20354e-01)
rgb = (9.72590e-01, 5.29798e-01, 5.33240e-02)
rgb = (9.86964e-01, 6.30485e-01, 3.09080e-02)
rgb = (9.84865e-01, 7.28427e-01, 1.20785e-01)
rgb = (9.66243e-01, 8.36191e-01, 2.61534e-01)
rgb = (9.46392e-01, 9.30761e-01, 4.42367e-01)
rgb = (9.88362e-01, 9.98364e-01, 6.44924e-01)
}}

\pgfplotsset{%
	marginlabels/.style={%
		ticklabel style={font=\fontsize{7pt}{8pt}\selectfont},
	},
	marginplot/.style={%
		height=.9\marginparwidth,
		width=.9\marginparwidth,
		scale only axis,
		axis y line=center,
		axis x line=middle,
		marginlabels,
	},
	cycle list={[indices of colormap={0,4,8,12,17} of flare]},
	pogmdm group plot/.style={%
		group/x descriptions at=edge bottom,
		group/y descriptions at=edge left,
		group/vertical sep=2.5mm,
		group/horizontal sep=2.5mm,
		width=1cm,
		height=1cm,
		scale only axis,
		no markers,
		ticklabel style={font=\tiny},
		grid=major,
		cycle list={[indices of colormap={0,4,8,12,17} of flare]},
		semithick,
	},
}
\def\wwidth{2.25cm}

\pgfmathsetlengthmacro\spysize{\wwidth / 2 - 4 * 0.2pt}
\tikzset{
	mrispy/.style={%
		spy using outlines={%
			rectangle,
			magnification=3,
			width=\spysize,
			height=\spysize,
            color=orange
		}
	},
	font=\footnotesize,
}

\newacronym{acl}{ACL}{auto-calibration line}
\newacronym{cg}{CG}{conjugate gradient}
\newacronym{cnn}{CNN}{convolutional neural network}
\newacronym{corpd}{CORPD}{coronal proton density}
\newacronym{corpdfs}{CORPDFS}{coronal proton density fat suppressed}
\newacronym{cs}{CS}{compressed sensing}
\newacronym{djs}{DJS}{deep-JSENSE}
\newacronym{dps}{DPS}{diffusion posterior sampling}
\newacronym{ebm}{EBM}{energy-based model}
\newacronym{elu}{ELU}{exponential linear unit}
\newacronym{ema}{EMA}{exponential moving average}
\newacronym{foe}{FOE}{fields of expert}
\newacronym{gan}{GAN}{generative adversarial network}
\newacronym{gmm}{GMM}{Gaussian mixture model}
\newacronym{grappa}{GRAPPA}{generalized autocalibrating partial parallel acquisition}
\newacronym{id}{ID}{in distribution}
\newacronym{map}{MAP}{maximum a posteriori}
\newacronym{mcmc}{MCMC}{Markov chain Monte Carlo}
\newacronym{mlp}{MLP}{multi layer perceptron}
\newacronym{mmse}{MMSE}{minimum mean squared error}
\newacronym{mri}{MRI}{magnetic resonance imaging}
\newacronym{nmse}{NMSE}{normalized mean squared error}
\newacronym{ood}{OOD}{out of distribution}
\newacronym{palm}{PALM}{proximal alternating linearized minimization}
\newacronym{pc}{PC}{predictor corrector}
\newacronym{pde}{PDE}{partial differential equation}
\newacronym{pi}{PI}{parallel imaging}
\newacronym{pogmdm}{PoGMDM}{product-of-Gaussian-mixture diffusion model}
\newacronym{psnr}{PSNR}{peak signal-to-noise ratio}
\newacronym{rss}{RSS}{root sum of squares}
\newacronym{sde}{SDE}{stochastic differential equation}
\newacronym{sense}{SENSE}{sensitivity encoding}
\newacronym{smash}{SMASH}{simultaneous acquisition of spatial harmonics}
\newacronym{ssim}{SSIM}{structural similarity}
\newacronym{tv}{TV}{total variation}
\newacronym{tgv}{TGV}{total generalized variation}
\newacronym{vn}{VN}{variational network}
\newacronym{zf}{ZF}{zero filled}

\title[Article Title]{Product-of-Gaussian-Mixture Diffusion Models for Joint Nonlinear MRI Reconstruction}
\author*[1]{\fnm{Laurenz} \sur{Nagler}}\email{lnagler@tugraz.at}
\author[2]{\fnm{Martin} \sur{Zach}}\email{martin.zach@epfl.ch}
\author[1]{\fnm{Thomas} \sur{Pock}}\email{thomas.pock@tugraz.at}
\affil*[1]{\orgdiv{Graz University of Technology}, \orgname{Institute of Visual Computing}, \postcode{8010} \state{Graz}, \country{Austria}}
\affil[2]{\orgdiv{École Polytechnique Fédérale de Lausanne}, \orgname{Biomedical Imaging Group and Center for Biomedical Imaging}, \postcode{1015} \state{Lausanne}, \country{Switzerland}}

\abstract{
Recently, diffusion models have attracted considerable attention for magnetic resonance image reconstruction due to their high sample quality. However, most existing methods rely on large networks with opaque time-conditioning mechanisms, and require offline coil sensitivity estimation. This results in limited interpretability of the reconstruction process and reduced flexibility in the acquisition setup.
To address these limitations, we jointly reconstruct the image and the coil sensitivities by combining the parameter-efficient product-of-Gaussian-mixture diffusion model as an image prior with a classical smoothness prior on the coil sensitivities.
The proposed method is fast and robust to both contrast and anatomical distribution shifts as well as changing k-space trajectories.
Finally, we propose a more expressive parameterization of the image prior which improves results in denoising and magnetic resonance image reconstruction.
}
\begin{document}
\maketitle
\section{Introduction}
\label{sec:introduction}
\Gls{mri} is a non-invasive imaging modality that is widely used in clinical practice. It 
provides high spatial resolution, versatile contrast mechanisms, and radiation-free operation. 
However, it typically requires long scan times which limits patient throughput and makes it prone to motion artifacts.

In recent years, numerous techniques that aim to reduce \gls{mri} acquisition time while preserving the diagnostic quality of the reconstructions~\cite{pruessmann1999sense,griswold2002grappa,block2007CSTV,knoll2011tgvformri,hammernik2018variationalnetworks,zach2023stabledeepmri} have been developed. As an example, \gls{pi} methods exploit complementary information from multiple receiver coils to reconstruct images from partially-observed Fourier-domain (k-space) measurements~\cite{deshmane2012parallelimaging}.
\Gls{pi} methods can be divided into two main categories. Auto-calibration methods—such as \gls{grappa}~\cite{griswold2002grappa}—estimate missing data from \glspl*{acl} in the k-space center to directly combine coil measurements. In contrast, image-domain methods—such as \gls{sense}~\cite{pruessmann1999sense}—rely on spatially varying coil sensitivities to combine the data in image space. The \gls{sense} formulation leads to a nonlinear inverse problem (described in detail in~\Cref{sec:methods}) due to the unknown coil sensitivities.

Coil sensitivity estimation is traditionally addressed by: (i) Offline estimation of the coil sensitivities using an additional low-resolution prescan~\cite{hamilton2017parallelimagingmri}, followed by smoothing via low-pass filtering, locally fitting low-order polynomial~\cite{pruessmann1999sense,irfan2016sensmap_estimation} or splines~\cite{ballester2001robustspline}. (ii) Post-processing estimation from the acquired data, including wavelet-based approaches~\cite{lin2001waveletsensest, lin2003waveletbasedsensitivityestimation} and region growing with subsequent polynomial fitting~\cite{ling2014iterativecoilsensitivity_regiongrowing}. One of the most popular methods in this category is ESPIRiT~\cite{uecker2014espirit}, which formulates coil sensitivity estimation as an eigenvalue problem for an operator derived from the~\gls{acl} region of the data. The eigenvectors associated with the second-largest eigenvalue of this operator are then the estimates of the coil sensitivities. (iii) Joint reconstruction of both the image and the coil sensitivities. The authors of~\cite{ying2007jsense} parameterize the coil sensitivities with low-order polynomials and 
simultaneously optimize the image and coefficients of the polynomials. The authors of~\cite{bauer2007nlininv} and~\cite{uecker2008nlininv} solve the nonlinear problem using an iteratively regularized Gauss--Newton method with an L2 penalty on the image. In~\cite{bauer2007nlininv}, the coil sensitivities are parameterized with a small number of basis functions, whereas~\cite{uecker2008nlininv} augments the objective function with classical penalties that promote smoothness. These works were extended in~\cite{knoll2012nlinv_var_penalties, uecker2007nlinvL1wavelet} to incorporate variational penalties such as~\gls{tv},~\gls{tgv}\cite{bredies2010tgv} or L1-wavelet on the image. Finally, in~\cite{majumdar2012isense} the authors propose an alternating reconstruction method that exploits low-rank structure of the coil sensitivities. 
\linebreak
The methods in (i) and (ii) require additional time, either for acquiring the prescan or for post-processing the acquired data thus reducing flexibility. Joint reconstruction approaches in (iii) offer greater flexibility, but they are often computationally demanding and typically rely on hand-crafted regularization terms. In contrast, our method retains the flexibility of a joint-reconstruction framework while replacing the hand-crafted regularization of the image with a data-driven prior.

Recently, data-driven reconstruction methods have attracted considerable attention in the~\gls{mri} community. In early approaches researchers modeled the map from the measured data to the reconstructed image with a neural network that was trained on a large dataset of such measurement-image pairs. Examples include pre-processing approaches that predict missing k-space data~\cite{han2020kspacedeeplearning}, post-processing approaches that refine a naive reconstruction~\cite{zbontar2018fastMRI}, and learned iterative schemes~\cite{hammernik2018variationalnetworks, aggarwal2017MoDL, arvinte2021deepjsense}. All of these methods have in common that they incorporate the forward model in their training. The forward model accounts for coil sensitivities, k-space trajectories and noise. Consequently, the performance of such methods typically deteriorates if in the data acquisition changes; for instance if a different k-space trajectory is used (we demonstrate this in~\Cref{ssec:simulation_study,ssec:parallel_imaging}). In more recent approaches, methods that explicitly separate likelihood and prior became popular. These are naturally more robust to such variations. In the context of MRI, this also removes the need for difficult-to-obtain fully-sampled multi-coil training data, since the prior can be learned from readily available DICOM images. Methods that follow this separation often leverage learned generative models such as generative adversarial networks~\cite{narnhofer2019invGANS}, diffusion models~\cite{jalal2021robustCS,chung2022scoreMRI,luo2023bayesianmri}, or models defined by a Gibbs distribution (often called 
energy-based models)~\cite{zach2023stabledeepmri, guan2023ebmmri}.

Among these, diffusion models have emerged as particularly powerful for \gls{mri} reconstruction and are currently state-of-the-art.
Nonetheless, several limitations remain: U-Net–type networks~\cite{ronneberger2015UNET} typically approximate the score\footnote{The gradient of the logarithm of the probability density of interest; see~\Cref{ssec:background_diffusion}.} of the target distribution directly but are not explicitly constrained to form a conservative vector field. Consequently, theoretical conditions required for gradient-based \gls{mcmc} methods to generate samples according to the model distribution are not satisfied~\cite{salimans2021shouldebmmodelscore}. Additionally, the time conditioning of these networks---often implemented via Fourier features or positional embeddings~\cite{tancik2020fourierfeatures,vaswani2017attentionisallyouneed}---has no explicit link to the underlying~\gls{sde} that governs the diffusion process.
Finally, models with millions of parameters often entail substantial inference times and often require heuristics to handle data 
of arbitrary size~\cite{chung2022scoreMRI,erlacher2023jointnonlinearmriinversion}.

To address the limitations of typical networks used for diffusion models, Zach et al.~\cite{zach2024POGMDM} recently proposed the \gls{pogmdm}, a diffusion model with a fields-of-experts–type architecture~\cite{roth2005fieldsofexperts}. The model is parameter-efficient, its time-conditioning mechanism is linked to the underlying~\gls{sde}, and the derived score forms a conservative vector field. In this work, we utilise a \gls{pogmdm} in a joint-reconstructing framework, thus eliminating the need for offline coil sensitivity estimation. Furthermore, we explore various possible extensions to the original \gls{pogmdm} architecture and evaluate their performance in both denoising and \gls{pi} reconstruction.

\subsection{Related Work}
Most existing methods for \gls{pi} reconstruction assume knowledge of the coil sensitivities~\cite{jalal2021robustCS,luo2023bayesianmri,ozturkler2023regularization, hu2024adobiadaptivediffusionbridge,aggarwal2017MoDL}, which are typically obtained via offline estimation using algorithms such as ESPIRiT~\cite{uecker2014espirit}. However, this approach is sensitive to patient motion, requires a calibration region at the center of k-space, and involves an additional gridding step for non-Cartesian k-space trajectories. Alternatively, Chung et al.~\cite{chung2022scoreMRI} avoid explicit coil-sensitivity estimation by reconstructing individual coil images and combining them into the final reconstruction through a pixel-wise \gls{rss} reduction.
While this bypasses the need for the estimation of the coil sensitivities, the pixel-wise~\gls{rss} reduction of individual coil images is known to yield biased reconstructions even in the noise-free case~\cite{larsson2003snroptimrssrec}. In addition, the computational cost of their approach grows linearly with the number of coils. The approach proposed in~\cite{chung2023parallel} addresses the problem of unknown coil sensitivities by a joint reconstruction aided by an additional diffusion prior on the coil sensitivities. Like the approach by Chung et al.~\cite{chung2022scoreMRI}, the computational complexity of this approach scales linearly with the number of coils and adds the overhead of training and tuning a second diffusion model. Similarly, in~\cite{lip2024proxdiffmri}, separate diffusion priors for the image and coil sensitivities are used as substitutes for proximal operators within a~\gls{palm} framework~\cite{Bolte2014palm}. While this enables joint updates of both components, it again necessitates training an additional diffusion prior. Furthermore, the authors employ generatively learned models in \gls{map}--based inference, which is known to be suboptimal~\cite{schmidt2010generativeperspective}. Murata et al.~\cite{murata2023gibbsddrm} tackle the same problem by combining a diffusion prior for the image with a simple hand-crafted prior on the coil sensitivities. However, their approach relies on an efficient computation of the singular value decomposition of the forward operator, which is computationally infeasible for undersampled \gls{pi}. 

Our method also performs a joint reconstruction of the image and coil sensitivities, but differs in its modeling and computational requirements. Instead of learning a separate diffusion model for the sensitivities~\cite{chung2023parallel,lip2024proxdiffmri}, we employ a simple smoothness prior and update the image and coil sensitivities in an alternating fashion. This avoids training an additional diffusion model and maintains scalability with respect to the number of coils. Moreover, unlike~\cite{murata2023gibbsddrm}, our approach does not require the computation of a singular value decomposition of the forward operator.

For more general surveys of diffusion models in inverse problems we refer the reader to~\cite{chung2025diffusionmodelsinverseproblems,daras2024surveydiffusionmodelsinverse}.

\subsection{Contributions}
This work constitutes an extension to our preliminary results that were published in the conference paper~\cite{nagler2025pogmdmmri}. There, we combine a \gls{pogmdm} as a prior on the images and a simple smoothness prior on the coil sensitivities in a joint-reconstruction framework.

In our preliminary work, we evaluated the proposed joint-reconstruction framework on \gls{corpd}-weighted knee images and assessed its robustness to shifts in the contrast mechanism using \gls{corpdfs}-weighted knee images. 
In the present work, we substantially extend this study and make the following contributions:
\begin{itemize}
	\item We broaden the evaluation protocol to explicitly account for anatomy- and contrast mechanism shifts by considering T1- and T2-weighted brain scans.
	\item We explore various possible extensions to the \gls{pogmdm} architecture, and conduct extensive numerical experiments to compare their performance in both denoising and \gls{pi} reconstruction.
\end{itemize}

The remainder of the paper is organized as follows. We provide background on inverse problems, their Bayesian resolution, diffusion models and their application to inverse problems in~\Cref{sec:background}. We introduce the proposed joint-reconstruction algorithm and discuss alternative parameterizations of the~\gls{pogmdm} used within the algorithm in \Cref{sec:methods}.
In \Cref{sec:results}, we present extensive numerical results and compare the different~\gls{pogmdm} parameterizations on in-distribution data and out-of-distribution data in the form of anatomy- and contrast-mechanism shifts. Finally, we conclude the paper and discuss future research directions in~\Cref{sec:conclusion}.

\section{Background}%
\label{sec:background}
\subsection{Inverse Problems}%
\label{ssec:inv_prob}
In inverse problems, the goal is to estimate an unknown signal \( \mathbf{x} \in \mathbb{C}^d \) from the measurements 
\begin{equation}
\label{eq:inv_prob_setup}
\mathbf{y} = \mathbf{A}(\mathbf{x}) + \mathbf{n}
\end{equation}
where the \emph{forward operator} \( \mathbf{A}: \mathbb{C}^d \rightarrow \mathbb{C}^m\) models the measurement process and \( \mathbf{n} \in \mathbb{C}^m \) is additive measurement noise.
The forward operator \( \mathbf{A} \) may be linear or nonlinear depending on the physics of the acquisition. For example, in the \gls{sense} model for \gls{pi} ---where the image and the coil sensitivities are unknown---\( \mathbf{A} \) is nonlinear (see~\Cref{ssec:img_recon_algo} for details).
In the Bayesian resolution of this problem, the unknown signal is modeled as a random variable, denoted \( \mathbf{X} \), with associated distribution \( p_\mathbf{X} \), referred to as the \emph{prior}. Given a signal \( \mathbf{x} \), the \emph{likelihood} \( p_{\mathbf{Y}|\mathbf{X}}(\,\cdot\,\mid \mathbf{x}) \) describes the probability of observing any measurement, given the signal is \( \mathbf{x} \). The likelihood is fully specified by the model of the measurement process and the model of the noise. Throughout this manuscript, we assume that the noise is Gaussian~\cite{gudbjartsson1995noise_in_mri}, which implies that the likelihood is given by
\begin{equation}
\label{eq:noise_model}
p_{\mathbf{Y}|\mathbf{X}}(\mathbf{y}\mid\mathbf{x}) \propto \exp \left(-\frac{1}{2\sigma_\text{n}^2} \norm{\mathbf{A}(\mathbf{x}) - \mathbf{y}}^2\right),
\end{equation}
where \(\sigma_\text{n}^2\) is the variance of the noise.
We can combine the likelihood and the prior via Bayes' rule, which states that the object of interest, the \emph{posterior} \( p_{\mathbf{X}|\mathbf{Y}} \), is given by
\begin{equation}
\label{eq:bayes_rule}
p_{\mathbf{X}|\mathbf{Y}}(\mathbf{x}\mid \mathbf{y}) \propto  p_{\mathbf{Y}|\mathbf{X}}(\mathbf{y}\mid \mathbf{x})p_{\mathbf{X}}(\mathbf{x}).
\end{equation}
The resolution of the inverse problem is then to analyze the posterior distribution for a fixed prior,
for instance through the computation of Bayes estimators such as the \gls{map} estimator (the mode of the posterior distribution) or the \gls{mmse} estimator (the expectation of the posterior distribution)~\cite{pereyra2019revisitingmap}.
Furthermore, the Bayesian approach to inverse problems enables uncertainty quantification of reconstructed signals.
More background on Bayesian inverse problems can be found in~\cite{Dashti2017}.
Diffusion models are a popular choice for data-driven priors due to their ability to capture complex, high-dimensional distributions, their high-visual-quality samples, and their empirically-demonstrated generalization capabilities~\cite{glaszner2025biggersintalways}.

\subsection{Diffusion Models}%
\label{ssec:background_diffusion}
In the previous section we showed that the modeling burden in the Bayesian resolution of inverse problems lies with the prior \( p_\mathbf{X} \).
Modeling that prior from a finite set of samples is extremely challenging in the high-dimensional problems that are encountered in imaging tasks.
A fruitful approach is to smooth the empirical distribution defined by its samples by a convolution with a Gaussian.
This makes the distribution more regular and thereby eases the estimation task~\cite{song2019genmodelling, habring2025energybasedmodelsinverseimaging}.
However, the choice of the variance of the Gaussian is not obvious:
When it is too small, the distribution is mostly unchanged and remains irregular and challenging to estimate.
When it is too large, important features may be lost.
To overcome this, Song et al.~\cite{song2019genmodelling} propose to smooth the density at a sequence of noise scales, and to learn a model conditioned on those. Approximate sampling from their model is achieved by using the unadjusted Langevin algorithm while gradually annealing the noise. In a follow-up work~\cite{song2021scorebased}, Song et al. generalize this idea
to \glspl*{sde} by letting the number of noise scales go to infinity. A general form of the recovered \emph{diffusion process} is given by
\begin{equation}
\label{eq:diffusion_process_sde}
	\mathrm{d}\mathbf{X}_t = \mathbf{f}(\mathbf{X}_t, t)\mathrm{d}t + g(t)\mathrm{d}\mathbf{W}_t
\end{equation}
with the initial condition \( \mathbf{X}_0 = \mathbf{X} \), where \( \mathbf{W}_t \) is the standard Wiener process~\cite{saerkkae2019sdebook}, \( \mathbf{f}: \mathbb{R}^d \times \mathbb{R}_+ \rightarrow \mathbb{R}^d\) is the \emph{drift term} and \( g: \mathbb{R}_+ \rightarrow \mathbb{R} \) the \emph{diffusion term}. In this work, we consider a variance-exploding \gls{sde}~\cite{song2021scorebased} that is defined by the choice \( \mathbf{f}(\mathbf{X}_t, t) = \mathbf{0} \) and \( g(t) = \sqrt{2} \), such that
\begin{equation}
\label{eq:diffusion_process}
    \mathrm{d}\mathbf{X}_t = \sqrt{2}\mathrm{d}\mathbf{W}_t.
\end{equation}
The Fokker-Planck equation~\cite{Risken1996} links the evolution of the random variable \( \mathbf{X}_t \) to the evolution of its associated density \( p_{\mathbf{X}_t} \).
Specifically, under our choice of the diffusion term the \gls{pde} that governs the evolution of \( p_{\mathbf{X}_t} \) leads to the standard heat equation with unit diffusivity constant \( \partial p_{\mathbf{X}_t}/\partial t = \Delta p_{\mathbf{X}_t}$, subject to the boundary condition $p_{\mathbf{X}_0} = p_\mathbf{X} \), where \( \Delta \) denotes the Laplace operator~\cite{saerkkae2019sdebook}. 
This \gls{pde} with the specified boundary condition admits the solution 
\begin{equation}
	\label{eq:solution_heat_equation}
	p_{\mathbf{X}_t} = N_{0, 2t\mathbf{I}} * p_\mathbf{X},
\end{equation}
where \(N_{\boldsymbol{\mu}, \mathbf{\Sigma}}(\mathbf{x}) = (2\pi)^{\frac{d}{2}}|\mathbf{\Sigma}|^{-\frac{1}{2}}\exp\big(-\tfrac{1}{2}\norm{\mathbf{x} - \boldsymbol{\mu}}_{\mathbf{\Sigma}^{-1}}^2 \big)\), \textit{i.e.} the convolution of the reference density with a Gaussian of variance $2t$~\cite{yanovsky2005pde}. Here, \( p_{\mathbf{X}_t} \) approaches an isotropic Gaussian distribution with variance \(2t\)  as \(t\) approaches infinity, regardless of the initial density \( p_\mathbf{X} \)~\cite[Theorem~2.1]{vazquez2017asymptotic_heat_equation}.

To sample from the prior distribution \( p_\mathbf{X} \), we must reverse this process and move from \( p_{\mathbf{X}_{T\rightarrow\infty}} \) to the data distribution \( p_{\mathbf{X}_0} \).
By Anderson's theorem~\cite{anderson1982reversetimediffusion}, the reverse-time diffusion process that reproduces the marginals of~\eqref{eq:diffusion_process_sde} is given by 
\begin{equation}
	\mathrm{d}\mathbf{X}_t = \left(\mathbf{f}(\mathbf{X}_t, t)- g(t)^2\nabla\log p_{\mathbf{X}_t}(\mathbf{X}_t)\right)\mathrm{d}t + g(t)\mathrm{d}\mathbf{W}_t,
\end{equation}
where \( \mathrm{d}t \) is an infinitesimally-small negative time step. Plugging in our choices for \( \mathbf{f} \) and \( g \) yields
\begin{equation}
	\label{eq:rev_diffusion}
	\mathrm{d}\mathbf{X}_t = \sqrt{2}\mathrm{d}\mathbf{W}_t - 2\nabla\log p_{\mathbf{X}_t}(\mathbf{X}_t)\mathrm{d}t.
\end{equation}

The only unknown quantity in~\eqref{eq:rev_diffusion} is the gradient of the logarithm of \( p_{\mathbf{X}_t} \)---\emph{the score}---for all \( t > 0 \). In practice, the score is approximated by the gradient of the logarithm of a parametric density \( p_\theta \) (with parameters \(\theta\)) or by directly parameterizing the vector field itself. 

For parameter identification, the close relation
\begin{equation}
\label{eq:tweedies_formula}
	\mathbb{E}[\mathbf{X}_0 \mid \mathbf{X}_t = \mathbf{x}] = \mathbf{x} + 2t\nabla \log p_{\mathbf{X}_t}(\mathbf{x}),
\end{equation}
between the score of \(p_{\mathbf{X}_t}\) and the conditional expectation of \(\mathbf{X}_0\) given that \(\mathbf{X}_t\) takes on a certain value, known as \emph{Tweedie's formula}~\cite{efron2011tweedie,robbins1956empiricalbayes} is crucial. We provide a full derivation in~\Cref{sec:appendix_tweedie}. This relationship provides a natural way to estimate the parameters \( \theta \in \Theta\), where \(\Theta\) denotes a possibly constrained parameter set, of a model \( p_\theta(\,\cdot\,, t) \approx p_{\mathbf{X}_t}\) via an \gls{mmse} denoising objective for all times \( t > 0\),
\begin{equation}
\label{eq:dns_objective}
    \min_{\theta \in \Theta} \int_0^{\infty} \mathbb{E}_{(\mathbf{x}_0,\mathbf{x}_t) \sim p_{\mathbf{X}_0,\mathbf{X}_t}}\left[\norm{\mathbf{x}_0 - \mathbf{x}_t - 2t\nabla \log p_{\theta}(\mathbf{x}_t,t)}^2\right] \mathrm{d}t.
\end{equation}
This objective is commonly referred to as \emph{denoising score matching}~\cite{vincent2011dsm,song2021scorebased} (see~\cite[Section 3.1]{meng2021Estimatinghigherordergradients} for a derivation from Tweedie's formula). Sampling from the joint distribution \( p_{\mathbf{X}_0,\mathbf{X}_t} \) at time \(	t \) is achieved via \emph{ancestral sampling}: The joint distribution \( p_{\mathbf{X}_0,\mathbf{X}_t} \) factorizes as \(p_{\mathbf{X}_t|\mathbf{X}_0}p_{\mathbf{X}_0} \), where drawing a sample \( \mathbf{x}_0 \sim p_{\mathbf{X}_0} \) amounts to selecting a random training example. A sample \( \mathbf{x}_t \sim p_{\mathbf{X}_t|\mathbf{X}_0=\mathbf{x}_0} \) is then constructed by adding appropriately scaled noise \( \mathbf{x}_t = \mathbf{x}_0 + \sqrt{2t}\mathbf{z} \) with \( \mathbf{z} \sim N_{0,\mathbf{I}} \) to the training example.

Given access to the learned approximate score \(\nabla \log p_\theta\), one can discretize the reverse \gls{sde}~\eqref{eq:rev_diffusion}, \textit{e.g.} by using the Euler–Maruyama scheme~\cite{saerkkae2019sdebook}, and approximately sample from the learned density. In practice, however, the reverse scheme is problematic as no limiting marginal \(\mathbf{X}_\infty\) exists. A common workaround is to set \( p_{\mathbf{X}_T} = N_{0, \sigma_{\text{max}\mathbf{I}}} \), where \( \sigma_{\text{max}} \) is the highest noise level in the diffusion process.

\subsection{Inverse Problems with Diffusion Priors}%
\label{ssec:inv_prob_diffusion}
To solve inverse problems in the diffusion framework, the reverse diffusion process stated in~\eqref{eq:rev_diffusion} has to be conditioned on the acquired measurements. Accordingly, the score of the distribution \( p_{\mathbf{X}_t} \) is replaced by score of the conditional distribution \( p_{\mathbf{X}_t|\mathbf{Y}} \). In detail, the reverse-time \gls{sde} then is

\begin{equation}
    \mathrm{d}\mathbf{X}_t = \sqrt{2}\mathrm{d}\mathbf{W}_t - 2\nabla\log p_{\mathbf{X}_t\mid \mathbf{Y}}(\mathbf{X}_t\mid \mathbf{y})\mathrm{d}t,
\end{equation}
with a suitable initial condition for \( \mathbf{X}_T \). Using Bayes' rule, this can be equivalently expressed as
\begin{equation}
	\label{eq:cond_rev_diff}
    \mathrm{d}\mathbf{X}_t = \sqrt{2}\mathrm{d}\mathbf{W}_t - 2\left (\nabla \log p_{\mathbf{Y}\mid \mathbf{X}_t}(\mathbf{y}\mid \mathbf{X}_t) + \nabla\log p_{\mathbf{X}_t}(\mathbf{X}_t)\right )\mathrm{d}t,
\end{equation}
where \( \nabla \log p_{\mathbf{Y}\mid \mathbf{X}_t} \) is the score of the time-dependent likelihood and \( p_{\mathbf{X}_t} \) the score of the time-dependent prior, which can be approximated by the score of the learned parametric distribution \( p_\theta(\,\cdot \,,t) \). However, the time-dependent likelihood of the measurement that can be written as
\begin{equation}
\label{eq:time_cond_likelihood_integral}
\int p_{\mathbf{Y}\mid \mathbf{X}_0}(\mathbf{y}\mid \mathbf{X}_0)p_{\mathbf{X}_0\mid \mathbf{X}_t}(\mathbf{X}_0 \mid \mathbf{X}_t)\mathrm{d}\mathbf{X}_0
\end{equation} 
is generally intractable in high-dimensional settings, as it requires marginalization over all possible \( \mathbf{X}_0 \).
Consequently, the likelihood term has to be approximated for all times \( t > 0\). In this work we use the simple approximation with the choice \(p_{\mathbf{X}_0\mid \mathbf{X}_t}(\mathbf{X}_0\mid \mathbf{X}_t) = \delta(\mathbf{X}_0 - \mathbf{X}_t)\) where \(\delta\) is the Dirac measure. The score of the likelihood for each \(t\) is then
\begin{equation}
\label{eq:time_cond_likelihood_integral_2}
\nabla \log p_{\mathbf{Y}\mid \mathbf{X}_t}(\mathbf{y}\mid \mathbf{X}_t) \approx \nabla \log p_{\mathbf{Y}\mid \mathbf{X}_0}(\mathbf{y} \mid \mathbf{X}_t).
\end{equation} 
Under the assumption of additive Gaussian noise, the time-conditional likelihood approximation becomes
\begin{equation}
	\nabla \log p_{\mathbf{Y}\mid \mathbf{X}_0=\mathbf{x}}(\mathbf{y}) \propto -\mathbf{J}_\mathbf{A}(\mathbf{x})^H(\mathbf{A}(\mathbf{x}) - \mathbf{y})
\end{equation}
where \( \mathbf{J}_{\mathbf{A}}(\mathbf{x}) \in \mathbb{C}^{d \times m} \) is the Jacobian of \(\mathbf{A}\) at \(\mathbf{x}\) and \(H\) denotes the Hermitian transpose. 
This approximation was proposed in~\cite{jalal2021robustCS} for linear \(\mathbf{A}\) in the context of~\gls{mri} reconstruction. Another popular alternative to this simple approximation was proposed in~\cite{chung2023diffusionposteriorsampling}. Here the authors choose \(p_{\mathbf{X}_0\mid \mathbf{X}_t}(\mathbf{X}_0\mid  \mathbf{X}_t) = \delta(\mathbf{X}_0 - \mathbb{E}[\mathbf{X}_0\mid \mathbf{X}_t])\), which leads to the approximation \( \nabla \log p_{\mathbf{Y}\mid \mathbf{X}_t}(\mathbf{X}_t\mid \mathbf{y}) \approx \nabla \log p_{\mathbf{Y}\mid \mathbf{X}_0}(\mathbf{y}\mid \mathbb{E}[\mathbf{X}_0\mid \mathbf{X}_t])\). The specific choices of \(p_{\mathbf{X}_0\mid \mathbf{X}_t}(\mathbf{X}_0\mid \mathbf{X}_t) \) and different approaches to the time-dependent likelihood approximation are subject to ongoing research. We refer to~\cite{daras2024surveydiffusionmodelsinverse} for a comprehensive overview.

\subsection{Product of Gaussian Mixture Diffusion Model}
\label{ssec:GMDiffusion}
In this section, we detail the~\gls{pogmdm}, describe its time adaptation mechanism, and explain why tractable diffusion of this model is unattainable in practice. The~\gls{pogmdm} models the time-dependent distribution of images \( \mathbf{x} \in \mathbb{R}^d \) as

\begin{equation}
\label{eq:model_eq}
    p_{\theta}(\mathbf{x}, t) \propto \prod_{l=1}^{d} \prod_{k=1}^o \psi_k((\mathbf{K}_k \mathbf{x})_{l}, \mathbf{w}_k, t),
\end{equation}
\( \mathbf{K}_1,...,\mathbf{K}_o \in \mathbb{R}^{d \times d}\) are convolution matrices\footnote{The convolution matrices \(\mathbf{K}_1, \dots, \mathbf{K}_o\) implement circular boundary conditions which ensure translation invariance of the prior.}
and the \(k\)-th factor
\begin{equation}
\label{eq:1d_gmm}
    \psi_k(x, \mathbf{w}, t) = \sum_{i=1}^L \frac{\mathbf{w}_{i}}{\sqrt{2\pi\sigma_k^2(t)}} \exp\left (-\frac{(x - \mu_i)^2}{2\sigma_k^2(t)} \right )
\end{equation}
is a one-dimensional Gaussian mixture with \(L\) components, weights $\mathbf{w} \in \mathbb{R}^L$, means \( \mu_1, \,...\, \mu_L \in \mathbb{R}\) that are shared across the potentials and variance \(\sigma_k^2 \in \mathbb{R_{++}}\) that is shared across the mixture components.

This model admits several advantages compared to the networks that typically serve as the backbone in diffusion models: the number of learnable parameters is small, the shallow architecture enables fast inference, the derived score is a conservative vector field, and, as we show later, the time conditioning adapts the variances \( \sigma_1^2,\dots,\sigma_o^2 \) based on the underlying \gls{sde}. However, these advantages come at the cost of limited model capacity and substantially inferior empirical performance compared to typical networks when used as priors in the resolution of inverse problems (see, \textit{e.g.}, \Cref{ssec:simulation_study}). The learnable parameters of the model detailed in~\eqref{eq:model_eq} and~\eqref{eq:1d_gmm} are the filters \( \mathbf{f}_k \in \mathbb{R}^r \) of the convolution matrices \( \mathbf{K}_k \) and the weights \(\mathbf{w}_k\) of the Gaussian mixture factors \(\psi_k\) for \(k=1,\dots,o\). These parameters are learned by minimizing the objective in~\eqref{eq:dns_objective}. All other parameters are chosen a priori and fixed (see \Cref{ssec:learning_details} for parameterization- and learning details).

Zach et al.~\cite{zach2024POGMDM} showed that if the convolution matrices \(\mathbf{K}_1,\dots,\mathbf{K}_o\) implement convolutions with circular boundary conditions and ideal filters\footnote{A filter is \emph{ideal} if it admits a constant gain in the passband, zero gain in the stopband and a linear phase response~\cite[Section 4.5.1]{proakis1996dsp}.} with non-overlapping supports, the convolution of \( p_\theta(\, \cdot \,,0) \) with \( N_{0, 2tI} \) in~\eqref{eq:solution_heat_equation} can be implemented solely by a proper adaption of the variances \(\sigma^2_1,\dots,\sigma^2_o\) of the one-dimensional Gaussian mixtures \(\psi_1,\dots,\psi_o\). In particular, if the variance of the \(k\)-th factor \( \psi_k \) is adapted as
\begin{equation}
\label{eq:1dgmm_var_adapt}
\sigma_k^2(t) = \sigma_0^2 + \nu_k^2 2t,
\end{equation}
where \( \sigma_0^2 \) is some initial variance (available as a modeling choice) and \( \nu_k \in \mathbb{R}\) is the magnitude spectrum of the \(k\)-th filter, then \(p_\theta(\, \cdot \,,t) = N_{0, 2t\mathbf{I}} * p_\theta(\, \cdot \,,0)\) for all \( t > 0 \). 

The proof of the time conditioning~\cite[Theorem~5]{zach2024POGMDM} relies on the assumption of ideal filters which is unattainable in practice as an ideal filters has infinite spatial support. Consequently, the evolution of the model distribution according to~\eqref{eq:diffusion_process} is infeasible.
As a practical compromise between filters with local spatial support and ideal one, Zach et al.~\cite{zach2024POGMDM} propose to use nonseparable shearlets~\cite{lim2013non_sep_shearlets}. They naturally partitions the frequency plane into cones determined by shearing and scaling parameters (see~\Cref{ssec:learning_details} for details). Nevertheless, they violate the ideal filter assumption. 

This naturally raises the question of which filters and corresponding time adaption of the variances are optimal in a practical sense. We attempt to answer this in~\Cref{ssec:alternative_pogmdm_parametrizations}, by allowing more general filters \(\mathbf{f}_k\) and parameterizations of the the map \(t \mapsto \sigma_k^2(t)\) for \(k=1,\dots,o\).

\section{Methods}
\label{sec:methods}
In this section, we introduce the acquisition model we use for \gls{pi}-\gls{mri}, specify our approximation for the time-dependent prior and likelihood, and discuss the employed model parameterizations.

\subsection{Acquisition Model and Reconstruction Algorithm}
\label{ssec:img_recon_algo}
We use the \gls{sense} acquisition model that relates the measured data \(\mathbf{y}\) to the unknowns by
\begin{equation}
\label{eq:non_lin_problem}
\mathbf{y} = \mathbf{A}(\mathbf{x},\mathbf{s}) + \mathbf{n} = 
    \begin{pmatrix}
        \mathbf{M}\mathbf{F}(\mathbf{s}_1 \odot \mathbf{x}) \\
        \vdots\\
        \mathbf{M}\mathbf{F}(\mathbf{s}_{c} \odot \mathbf{x})
    \end{pmatrix}
    + \mathbf{n}
\end{equation}
and jointly reconstruct the unknown image \( \mathbf{x} \in \mathbb{C}^{d} \) as well as the unknown coil sensitivities \( \mathbf{s} = (\mathbf{s}_1, \dots \mathbf{s}_c) \in \mathbb{C}^{d \times c} \) of \( c \in \mathbb{N} \) individual coils from the measured data \( \mathbf{y} = (\mathbf{y}_1, \dotsc, \mathbf{y}_{c}) \in \mathbb{C}^{e \times c} \) where \( e \in \mathbb{N} \) is the number of measured spatial frequencies. The nonlinear forward operator \(\mathbf{A}: \mathbb{C}^{d} \times \mathbb{C}^{d \times c} \to \mathbb{C}^{e \times c}\) models an element-wise multiplication between the image \(\mathbf{x}\) and the coil sensitivities \(\mathbf{s}_1,\dots,\mathbf{s}_c\), denoted as \( \odot \), followed by a discrete Fourier transform \(\mathbf{F}: \mathbb{C}^{d} \to \mathbb{C}^{d} \) and a binary sampling operator \( \mathbf{M} \in \mathbb{C}^{e \times d} \). The noise term \( \mathbf{n} \in \mathbb{C}^{e \times c} \) is modeled as additive complex Gaussian noise.

Recovering \( \mathbf{x} \) and \( \mathbf{s} \) from \( \mathbf{y} \) is ill-posed even in the fully sampled case (\textit{i.e.} when the sampling operator \(\mathbf{M}\) is the identity), due to a scaling ambiguity~\cite{uecker2008jointrecon}: for any \( \mathbf{b} \in \mathbb{C}^{d} \), if \( (\mathbf{x}^*, \mathbf{s}^*) \) is a solution to the recovery problem, then so is \( (\mathbf{x}^* \odot \mathbf{b}, \mathbf{s}^* \oslash \mathbf{b}) \), where the symbol \(\oslash\) denotes element-wise division. 

We address this problem in the diffusion framework, and introduce a time-dependent prior \(p_{\mathbf{X}_t, \mathbf{\Sigma}_t}\) on the joint distribution of \( \mathbf{X}_t \) and \( \mathbf{\Sigma}_t \). Here \( \mathbf{X}_t \) denotes the random variable of the image, and \( \mathbf{\Sigma}_t \) the random variable of the coil sensitivities, both of which indexed with the diffusion time \(t\). From the forward model in~\eqref{eq:non_lin_problem}, we derive the likelihood \( p_{\mathbf{Y}|\mathbf{X}_t,\mathbf{\Sigma}_t}(\mathbf{y} \mid \mathbf{x}, \mathbf{s}) \propto \exp\left(-\tfrac{1}{2}\norm{\mathbf{A}(\mathbf{x},\mathbf{s}) - \mathbf{y}}_2^2\right) \), where we make use of the approximation discussed in~\Cref{ssec:inv_prob_diffusion}. For a given datum \(\mathbf{Y} = \mathbf{y}\), we denote the time-dependent posterior as
\begin{equation}
\begin{aligned}
p_{\mathbf{X}_t,\mathbf{\Sigma}_t\mid \mathbf{Y}}(\mathbf{x}, \mathbf{s} \mid \mathbf{y})
&\propto p_{\mathbf{Y}|\mathbf{X}_t,\mathbf{\Sigma}_t}(\mathbf{y} \mid \mathbf{x}, \mathbf{s}) \, p_{\mathbf{X}_t,\mathbf{\Sigma}_t}(\mathbf{x}, \mathbf{s}) \nonumber\\
&= p_{\mathbf{Y}|\mathbf{X}_t,\mathbf{\Sigma}_t}(\mathbf{y} \mid \mathbf{x}, \mathbf{s}) \, p_{\mathbf{X}_t}(\mathbf{x}) \, p_{\mathbf{\Sigma}_t}(\mathbf{s}),
\label{eq:full_posterior}
\end{aligned}
\end{equation}
where we make use of Bayes' theorem and the simplifying assumption that the joint distribution of \( (\mathbf{X}_t, \mathbf{\Sigma}_t) \) factorizes. Although this factorization neglects correlations between coil sensitivities and the scanned anatomy, it has been observed to work well in practice~\cite{zach2023stabledeepmri,erlacher2023jointnonlinearmriinversion}.
For the time-dependent prior on the image \( p_{\mathbf{X}_t}\), we use the learned \gls{pogmdm} \( p_{\theta}(\, \cdot \,, t) \) trained on clinically available magnitude reference reconstructions. Note that this introduces a slight mismatch between the random variable \( \mathbf{X}_t \) which takes valued in \( \mathbb{C}^d \) and the prior that is learned on real-valued signals. To address this, we follow the heuristic proposed in~\cite{chung2022scoreMRI} and apply the prior separately to the real- and imaginary-parts of the image.
For the prior on the coil sensitivities, we follow~\cite{zach2023stabledeepmri} who choose a classical smoothness prior \( p_{\mathbf{\Sigma}_t}(\mathbf{s}) \propto \exp(-\gamma(\mathbf{s})) \), with
\begin{equation}
\label{eq:smoothness_prior}
    \gamma:\mathbb{C}^{d \times c} \rightarrow \mathbb{R}:
    \mathbf{s} \mapsto \frac{1}{2}\sum_{i=1}^{c} \left( \norm{\mathbf{D}\text{Re}(\mathbf{s}_{i})}_2^2 + \norm{\mathbf{D}\text{Im}(\mathbf{s}_{i})}_2^2\right),
\end{equation}
where \( \mathbf{D}:\mathbb{R}^{d} \rightarrow \mathbb{R}^{2d} \) is a forward-finite-differences operator with Dirichlet boundary conditions that ensure that the coil sensitivities are zero outside the image domain~\cite{zach2023stabledeepmri}.
For reconstruction, we follow~\cite{chung2022scoreMRI} and use the predictor--corrector algorithm introduced in~\cite{song2021scorebased} to solve the reverse~\gls{sde}. Here the predictor implements an Euler-Maruyama discretization of the reverse-time \gls{sde} and the corrector applies annealed Langevin dynamics~\cite{song2019genmodelling}. For the predictor we adopt the discretization schedule \( \zeta(t) = \zeta_{\mathrm{max}}(\zeta_{\mathrm{min}}/\zeta_{\mathrm{max}})^{(1 - t/T)^p} \) where \( t \in [0, T] \) and \( p \in \mathbb{R_+} \) (see appendix \Cref{sec:appendix_A} for more details). 
In each step of the reverse~\gls{sde}, data consistency is enforced by doing gradient descent steps, where the gradient of the log-likelihood w.r.t. \( \mathbf{x} \) for fixed coil sensitivities is 
\begin{equation}
\label{eq:grad_x_liklihood}
    \nabla_{\mathbf{x}} \log p_{\mathbf{Y}|\mathbf{X}_t, \mathbf{\Sigma}_t}(\mathbf{y}\mid \mathbf{x}, \mathbf{s}) \propto \sum_{i=1}^{c} \bar{\mathbf{s}}_{i} \odot (\mathbf{F}^H\mathbf{M}^H(\mathbf{y}_i - \mathbf{M}\mathbf{F}(\mathbf{s}_{i} \odot \mathbf{x}))), 
\end{equation}
where the overline denotes complex-conjugation. For updating the coil sensitivities, we found it sufficient to fix the prior for all \( t > 0 \). In this step, we depart from the standard diffusion framework and instead perform proximal-gradient updates on the sensitivities \( \mathbf{s} \). The proximal map of \( -\log p_{\mathbf{\Sigma}_t} \) admits the closed-form solution 
\begin{equation}
\text{prox}_{\mu \gamma}(\mathbf{s}) = 
\begin{pmatrix}
    Q_\mu(\text{Re}(\mathbf{s}_{1})) + iQ_\mu(\text{Im}(\mathbf{s}_{1})) \\
    \vdots \\
    Q_\mu(\text{Re}(\mathbf{s}_{c})) + iQ_\mu(\text{Im}(\mathbf{s}_{c}))
\end{pmatrix},
\label{eq:sens_prox}
\end{equation}
with \( \mu \in \mathbb{R}_{+} \) and \( Q_\mu:\mathbb{R}^{d} \rightarrow \mathbb{R}^{d}; \mathbf{x} \mapsto \mathbf{S}^T\left(\mathbf{S}(\mu \mathbf{x}) \odot (\boldsymbol{\tau} + \mu)^{-1}\right) \).
Here, \( \mathbf{S}: \mathbb{R}^{d \times d} \) is the two-dimensional discrete sine transform, and \( \boldsymbol{\tau} \) are the eigenvalues of the two-dimensional discrete Laplace operator~\cite{press1992numericsinc}. The application of this proximal operator in each step of the reverse diffusion ensures smoothness of the coil sensitivities.
The gradient of the log-likelihood w.r.t. to the coil sensitivities is 
\begin{equation}
\label{eq:grad_sigma_likelihood}
\nabla_{\mathbf{s}} \log p_{\mathbf{Y}|\mathbf{X}_t,\mathbf{\Sigma}_t}(\mathbf{y}\mid \mathbf{x}, \mathbf{s}) = 
\begin{pmatrix}
    \mathbf{x} \odot (\mathbf{F}^H\mathbf{M}^H(\mathbf{y}_1 - \mathbf{M}\mathbf{F}(\mathbf{s}_{1} \odot \mathbf{x}))) \\
    \vdots \\
    \mathbf{x} \odot (\mathbf{F}^H\mathbf{M}^H(\mathbf{y}_c - \mathbf{M}\mathbf{F}(\mathbf{s}_{c} \odot \mathbf{x})))
\end{pmatrix}.
\end{equation}

\begin{algorithm}
	\caption{Reconstruction algorithm for the resolution of~\eqref{eq:non_lin_problem}}\label{alg:joint_recon}
	\begin{algorithmic}[1]
		\Ensure \( \mathbf{x}_0, \mathbf{s}_0 \)
		\For{\( i = N-1,\,...\,,0 \)}
			\For{\( k = \textnormal{Re},\textnormal{Im} \)}\Comment{Predictor}
				\State \( \mathbf{x}_{k,i} \gets \mathbf{x}_{k,i+1} + (\zeta_{i+1}^2 - \zeta_{i}^2)\nabla\log p_{\theta}(\mathbf{x}_{k,i+1},\zeta^2_{i+1}) \)
				\State \( \boldsymbol{\xi}_i^{k} \sim N_{0,I} \)
				\State \( \mathbf{x}_{k,i} \gets \mathbf{x}_{k,i} + \sqrt{\zeta_{i+1}^2 - \zeta_{i}^2}\boldsymbol{\xi}_i^{k} \)
			\EndFor
			\State \( \mathbf{x}_{i} \gets \mathbf{x}_{\text{Re},i} + \text{i}\mathbf{x}_{\text{Im},i} \)
			\State \( \mathbf{x}_{i} \gets \mathbf{x}_{i} + \lambda\nabla_{\mathbf{x}_i}\log p_{\mathbf{Y}|\mathbf{X}_t,\mathbf{\Sigma}_t}(\mathbf{y} \mid \mathbf{x}_i, \mathbf{s}_{i+1}) \)\;
			\For{\( j = 1\,...\,M \)}\Comment{Corrector}
				\For{\( k = \textnormal{Re},\textnormal{Im} \)}
					\State \( \boldsymbol{\xi}_j^{k} \sim N_{0,I} \)
					\State \( \mathbf{x}_{k,i} \gets  \mathbf{x}_{k,i} + \epsilon_i \nabla\log p_{\theta}(\mathbf{x}_{k,i}, \zeta^2_{i}) + \sqrt{2\epsilon_i}\boldsymbol{\xi}_j^{k} \)\;
				\EndFor
				\State \( \mathbf{x}_{i} \gets \mathbf{x}_{\text{Re},i} + \text{i}\mathbf{x}_{\text{Im},i} \)
				\State \( \mathbf{x}_{i} \gets \mathbf{x}_{i} + \lambda\nabla_{\mathbf{x}_i}\log p_{\mathbf{Y}|\mathbf{X}_t,\mathbf{\Sigma}_t}(\mathbf{y}\mid \mathbf{x}_i, \mathbf{s}_{i+1}) \)\;
			\EndFor
			\State \( \mathbf{s}_i \gets \text{prox}_{\mu \gamma}\bigl(\mathbf{s}_{i+1} + \mu\nabla_{\mathbf{s}_{i+1}} \log p_{\mathbf{Y}|\mathbf{X}_t,\mathbf{\Sigma}_t}(\mathbf{y}\mid \mathbf{x}_i, \mathbf{s}_{i+1})\bigr) \)\Comment{Coil Update}
		\EndFor
	\end{algorithmic}
\end{algorithm}
We summarize the joint reconstruction in~\Cref{alg:joint_recon}, where \( \lambda \in \mathbb{R}_{\geq0} \) and \( \mu \in \mathbb{R}_{\geq0} \) are hyperparameters that balance the data consistency. The step size \( \epsilon \) at iteration \(i\) is \(2r\norm{\mathbf{\boldsymbol{\xi}}}_2^2 / \norm{\nabla \log p_\theta(\, \cdot \,, \zeta^2_{i})}_2^2 \) where \(\boldsymbol{\xi} \sim N_{0,I}\), \(\zeta\) the noise schedule at \(i\) and \( r \in \mathbb{R}_+\) is a hyperparameter. Finally, we want to stress that the resulting algorithm departs significantly from the resolution of the conditional reverse diffusion. Due to the coil sensitivity updates and approximate data consistency, the approach is closer to a stochastic graduated-nonconvexity heuristic than a rigorous probabilistic resolution of the problem.

\subsection{Extended \gls{pogmdm} Parameterization}
\label{ssec:alternative_pogmdm_parametrizations}
To balance theoretical assumptions with practical utility of the filters, the original \gls{pogmdm} formulation~\cite{zach2024POGMDM} employs nonseparable shearlets. In this work, we explore two alternative parameterizations:
(i) We parameterize the filters as fully-learnable while retaining the original time conditioning detailed in~\eqref{eq:1dgmm_var_adapt}, and
(ii) additionally learn the time conditioning directly from data.
As we cannot assume ideal filters, the scaling for the diffusion time \(\nu_k^2\) in~\eqref{eq:1dgmm_var_adapt} has to be calculated with a heuristic. Here, we follow~\cite{zach2024POGMDM} and set \( \nu_k^2 = \norm{|\mathbf{F}\mathbf{f}_k|}_\infty \). We also experimented with an alternative heuristic given by \( \nu_k^2 = \tfrac{1}{b}\sum_b(|\mathbf{F}\mathbf{f}_k|)_b\) but observed no significant difference.
In (ii) we adapt the variances of the \(k\)-th one-dimensional Gaussian mixture factor as
\begin{equation}
	\sigma^2_k(t) = \sigma_0^2 + (\tau_{\theta}(t))_k
	\label{eq:learned_embedding}
\end{equation}
with the diffusion time \( t \) for \(k = 1,\dots,o\). Here, \( \tau_\theta: \mathbb{R} \rightarrow \mathbb{R}^o_+ \) maps the time \( t \) to 
an \( o \)-dimensional positive vector. We parameterize \( \tau_\theta \) as the map
\( t \mapsto (\text{softplus} \circ L_3 \circ \text{ELU} \circ L_2 \circ \text{ELU} \circ L_1)(\sqrt{2t}) \),
with linear layers \( L_i: x \mapsto W_ix + b_i,\, i = 1,2,3\) and the \gls{elu} function with unit scaling. The final \( \text{softplus}: x \mapsto \ln(1 + \exp(x))\) ensures that the outputs are strictly positive.
By learning the time conditioning from data, the model partially compensates for empirical diffusion inaccuracies in the original \gls{pogmdm}. This approach also enables the dynamic disabling of filters with diffusion time \(t\) (see \Cref{ssec:learned_models} for a detailed discussion).

\section{Numerical Experiments}
In this section, we first describe the data used in our experiments. We then outline the experimental setup, including the parameter settings of the joint reconstruction algorithm and the evaluation metrics. Finally, we provide learning details for our used models.

\subsection{Experimental Data}%
\label{ssec:exp_data}
We use the fastMRI dataset~\cite{knoll2020fastmri} for training, hyperparameter search, and evaluation. 
The training data consists of root-sum-of-squares reconstructions of the central eleven slices of the \gls{corpd} training split to avoid training on noise-only data.
This yields 5324 training images of size $320 \times 320$ which we normalize with the map $\mathbf{x} \mapsto \mathbf{x} / \norm{\mathbf{x}}_{\infty}$.

The \gls{corpd} validation split is divided into 30 validation and 58 test files, excluding k-spaces with width different from \(368\) and \(372\) to balance dataset size and computational cost. To test the robustness of our reconstruction algorithm on contrast-mechanism out-of-distribution data, we conduct experiments with the \gls{corpdfs} knee validation split. We apply the same exclusion criteria as for the \gls{corpd} dataset, which results in 30 validation and 69 test files.
For anatomical out-of-distribution experiments, we use contrast-agent T1-and T2-weighted brain scans from the fastMRI brain dataset~\cite{knoll2020fastmri}. We exclude k-space data with 
width different from \( 320 \) and \(322\) and restrict our experiments to data with four coils to manage the computational cost of the experiments. This yields 20 validation and 43 test files for T1 and 20 validation and 52 test files for T2.

\subsection{Experimental Setup}%
\label{ssec:comp_eval_methods}
The reconstruction algorithm is evaluated in two experiments: (i) a real-valued single-coil simulation with synthetic data, and (ii) joint \gls{pi} reconstruction on both in-distribution (\gls{corpd}) and out-of-distribution (\gls{corpdfs}, T1 brain and T2 brain) datasets. Details for the experiments and the comparison methods are provided in the respective sections presenting the numerical results.

We also conduct a concise ablation study of the \gls{pogmdm} parameterizations for the shearlet model and the learned time conditioning. For all joint reconstruction experiments, we report \gls{psnr}, \gls{nmse}, and \gls{ssim} of the magnitude of the reconstructed image relative to the reference \gls{rss} reconstructions. Unless stated otherwise, we obtain and report Monte-Carlo estimates of the \gls{mmse} estimate of the diffusion posterior by averaging 25 independent reconstructions.

To ensure a fair comparison, our method includes an additional postprocessing step to match the intensity distribution of the reference \gls{rss} reconstructions. Following~\cite{uecker2008nlininv}, we correct small intensity shifts by multiplying the final reconstruction with the \gls{rss} of the estimated coil sensitivities:
\begin{equation}
	\mathbf{x} \mapsto \mathbf{x} \cdot \sqrt{\sum_{i=1}^{c} |\mathbf{s}_i|^2}.
\end{equation}
Similar to~\cite{zach2023stabledeepmri}, we fit a spline curve against the empirical scatter of reconstructed and reference intensities to adjust background intensities.

For all experiments, the parameters in~\Cref{alg:joint_recon} are set to \(N=1000\), \(M=1\) and \(\lambda=1\). We find an optimal step size parameter \(r\) and trade off parameter \(\mu\) for each k-space trajectory with grid search. For the noise schedule in~\eqref{eq:noise_schedule}, we set \(\zeta_{\mathrm{max}} = 10\), \(\zeta_{\mathrm{min}} = 0.001\), \(p=5\) and \(T=1\) in all experiments. Finally, we use the acceleration method proposed in~\cite{chung2022ccdf} with vanilla initialization and a reduced reverse diffusion starting time of \(t_0=0.2 T\).

\subsection{Learning Details}%
\label{ssec:learning_details}
In this section we provide parameterization- and learning details of the models used in our experiments.
\subsubsection{Filters Selection}
For the shearlet \gls{pogmdm}, we largely follow~\cite{zach2024POGMDM} and use nonseparable shearlet filters.
These filters are constructed by specifying, a scaling \( j > 0 \), translations \( m \in \mathbb{Z} \), and a range of 
shearings \( |k| \leq \left\lceil 2^{\lfloor j/2 \rfloor} \right\rceil \). 
The maximum number of shearings per scale is \( 2 |k| + 1 \). In all experiments, we use the maximal possible number of shearings per scale \textit{e.g.}, a scaling of \( j=2 \) leads to \( 5 \) shearings per scale. Considering both the vertical and horizontal cones, this yields \( o = 2 \times 5 \times 2 = 20 \).
The shearlet system itself is constructed from a one-dimensional lowpass filter 
\( \mathbf{h}_1 \) and a two-dimensional directional filter \( \mathbf{P} \), both of which are learned during training unless specified otherwise.
We use the same initialization for the filters as in~\cite{zach2024POGMDM} and ensure that \( \mathbf{h}_1 \) and \( \mathbf{P} \) define a valid shearlet system by projecting them onto suitable constraints during their iterative numerical optimization, see the details in~\cite{zach2024POGMDM}.

For the model whose filters \(\mathbf{f}_1,\dots,\mathbf{f}_o \in \mathbb{R}^{v \times v}\) are fully learnable we set \(v = 5\), and initialize the weights using the normal distribution \(N_{0, \frac{1}{ov^2}\mathbf{I}}\). Like in~\cite{chen2014insightintoanalysisop}, we enforce that the mean of the filters is zero to ensure invariance to radiometric shifts. This prevents the model from responding to global intensity biases often found in MRI images~\cite{Nyul1999onstandardizingMRIintensityscale}.

\subsubsection{Gaussian-Mixture Factors}
For the parameterization of the one-dimensional Gaussian mixture factors, we follow~\cite{zach2024POGMDM} and use \( L = 125 \) components whose means are spaced equidistantly over a predefined interval \([v_{\text{min}}, v_{\text{max}}]\). Specifically, we set \(v_{\text{min}}=-0.5\) and \(v_{\text{max}}=0.5\) for all shearlet models and the model with learned time conditioning. For the model with fully-learned filters and the original time conditioning, we found it benefitial to set \(v_{\text{min}}=-1\) and \(v_{\text{max}}=1\).
The mixture weights are constrained to the \( L \)-dimensional unit simplex and symmetry is
enforced by mirroring the first  \( \lceil L/2 \rceil - 1 \) entries of \( \mathbf{w}_k \) for \(k = 1,\dots,o\) around zero. Moreover, all Gaussian mixture components share a common initial variance \( \sigma_0^2 = (v_{\text{max}} - v_{\text{min}}) / (L -1) \). 

\subsubsection{Parameter Identification}
For parameter identification, we minimize the objective stated in~\eqref{eq:dns_objective} using the AdaBelief optimizer~\cite{zhuang2020adabelief}, complemented by projection steps to enforce the parameter constraints discussed above. Similar to~\cite{song2021scorebased}, we approximate the expectation over the time \( t \) in the interval \( [0, T] \) by sampling from a uniform distribution \( t \sim U_{0,T} \) with \( T=1 \). We train for \( 100\,000 \) iterations with parameter-group-specific learning rates depending on the chosen model parameterization. Furthermore, we apply an exponential moving average of the parameters with a momentum factor of \( 0.999 \).

\section{Results and Discussion}%
\label{sec:results}
We first visualize the potentials and the filters of the learned \glspl*{pogmdm}, and provide a detailed discussion of the variant which implements the learned time conditioning in \Cref{ssec:learned_models}. Then, in \Cref{ssec:simulation_study}, we report quantitative results for the real-valued single-coil simulation with synthetic data. Finally, we show numerical results for \gls{pi} on both in-distribution, contrast and anatomical out-of-distribution data in~\Cref{ssec:parallel_imaging}.

\subsection{Learned Models}%
\label{ssec:learned_models}
\subsubsection{Shearlets}
\begin{figure*}
	\centering
	\includegraphics{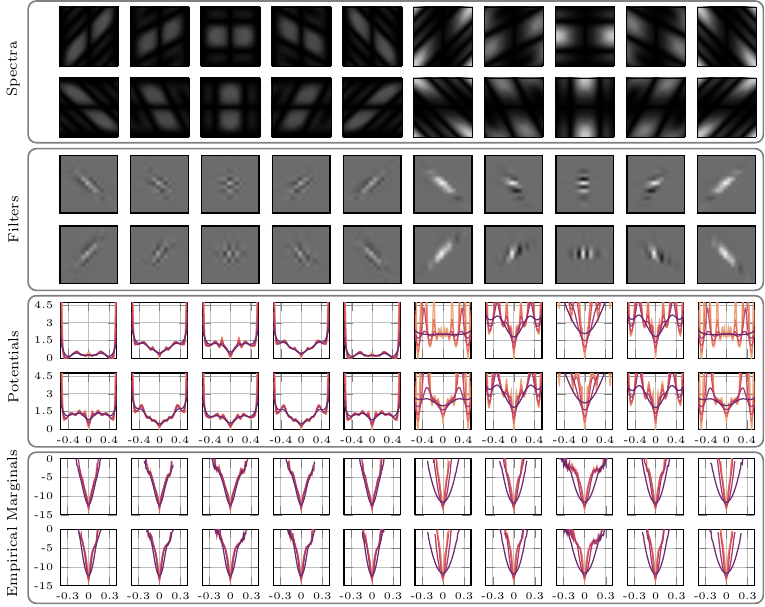}
    \caption{The learned shearlet model with Gaussian mixture factors and two shearlet scales. The first block shows the spectra of the learned shearlet filters. In the second block, the first row shows the vertical cone and the second row the horizontal cone of the shearlet system.
	The first five entries correspond to the shearing for the first scale and the second five for the second scale. The potentials and the negative logarithm of the empirical response marginals are shown for different diffusion times 
    $
    \sqrt{\num{2}t} = 
    \num{0} \protect\tikz[baseline=-\the\dimexpr\fontdimen22\textfont2\relax]\protect\draw [index of colormap={0} of flare, thick] (0,0) -- (.5, 0);,
    \num{0.025} \protect\tikz[baseline=-\the\dimexpr\fontdimen22\textfont2\relax]\protect\draw [index of colormap={4} of flare, thick] (0,0) -- (.5, 0);,
    \num{0.05} \protect\tikz[baseline=-\the\dimexpr\fontdimen22\textfont2\relax]\protect\draw [index of colormap={8} of flare, thick] (0,0) -- (.5, 0);,
    \num{0.1} \protect\tikz[baseline=-\the\dimexpr\fontdimen22\textfont2\relax]\protect\draw [index of colormap={12} of flare, thick] (0,0) -- (.5, 0);,
    \num{0.2} \protect\tikz[baseline=-\the\dimexpr\fontdimen22\textfont2\relax]\protect\draw [index of colormap={17} of flare, thick] (0,0) -- (.5, 0);
    $.}
    \label{fig:learned_model}
\end{figure*}

We show the learned filters implemented by the convolution matrices \( \mathbf{K}_1,\dots, \mathbf{K}_o \) and their corresponding potentials \(- \log\psi_1(\,\cdot\,, \mathbf{w}_1, t), \dots, - \log\psi_o(\,\cdot\,, \mathbf{w}_o, t) \) for different diffusion times in~\Cref{fig:learned_model}.
Due to the overcompleteness of the model---\textit{i.e.}, the fact that the number of filter responses \(d \cdot o\) exceeds the input dimension \(d\)---the learned potentials do not resemble the marginal distributions of the filter responses (see \textit{e.g.}, the empirical response marginals in~\Cref{fig:learned_model}). Instead, they are more complexly shaped and have multiple local minima, sometimes different from zero (\textit{e.g.}, the first and last potential in the vertical cone of the second scale in~\Cref{fig:learned_model}). This enables the amplification of specific structures in the images when employing this diffusion prior
\cite{zach2024POGMDM}. The authors of~\cite{zhu1997priorlearning,chen2017tnrd} report similar observations. 
Furthermore, the potentials associated with filters at scaling \( j=1 \) show smaller variations compared to those at \( j=2 \). This can be attributed to the higher magnitude maximum in the filter spectra, which in turn leads to stronger variation with the diffusion time (see~\eqref{eq:1dgmm_var_adapt}).
Similar to~\cite{zach2024POGMDM}, the spectra are overlapping and non-constant on their support, which shows that the theoretical assumptions discussed in~\Cref{ssec:GMDiffusion} are not met.

Interestingly, the learned potentials exhibit symmetries both within cones (\textit{e.g.}, the first and fifth potential in the first row) and across cones (\textit{e.g.}, potentials between rows) for a given scaling \( j \).
These symmetries likely arise from the structure of the training data which is evident in the empirical response marginals in~\Cref{fig:learned_model}. They suggest that the number of parameters in the model can be reduced without sacrificing its expressiveness. We present numerical results supporting this observation in~\Cref{ssec:ablation}. 

\subsubsection{Fully-learned Filters}
\begin{figure*}
	\centering
	\includegraphics{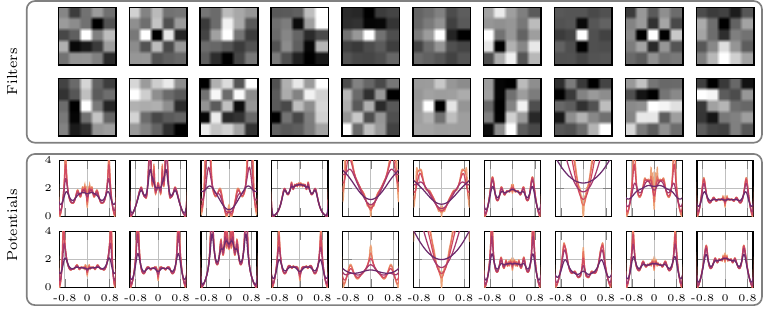}
    \caption{The learned model with Gaussian mixture factors and fully-learned zero-mean filters of size $(5 \times 5)$. The variances of the Gaussian mixture factors evolve with diffusion time according to \eqref{eq:1dgmm_var_adapt}. The potentials are shown for different diffusion times
		$
		\sqrt{\num{2}t} = 
		\num{0} \protect\tikz[baseline=-\the\dimexpr\fontdimen22\textfont2\relax]\protect\draw [index of colormap={0} of flare, thick] (0,0) -- (.5, 0);,
		\num{0.025} \protect\tikz[baseline=-\the\dimexpr\fontdimen22\textfont2\relax]\protect\draw [index of colormap={4} of flare, thick] (0,0) -- (.5, 0);,
		\num{0.05} \protect\tikz[baseline=-\the\dimexpr\fontdimen22\textfont2\relax]\protect\draw [index of colormap={8} of flare, thick] (0,0) -- (.5, 0);,
		\num{0.1} \protect\tikz[baseline=-\the\dimexpr\fontdimen22\textfont2\relax]\protect\draw [index of colormap={12} of flare, thick] (0,0) -- (.5, 0);,
		\num{0.2} \protect\tikz[baseline=-\the\dimexpr\fontdimen22\textfont2\relax]\protect\draw [index of colormap={17} of flare, thick] (0,0) -- (.5, 0);
		$.}
		\label{fig:learned_model_conv}
\end{figure*}

We show in \Cref{fig:learned_model_conv} the learned filters and their associated potentials. Some of the learned filters resemble classical ones: for example, the eighth filter in the first row and the sixth filter in the second row resemble second-order derivative filters in the vertical and horizontal directions.
The learned potentials again exhibit complex shapes due to overcompleteness. Interestingly, unlike in the shearlet model, some potentials show sharp peaks at zero (e.g., the fifth and eighth potential in the bottom row of~\Cref{fig:learned_model_conv}). Similar to the shearlet model, we observe that some of the learned potentials only show small variation with diffusion time, while others show stronger variations, which is again a consequence of the strong variation in the maximum of the magnitude spectra of the learned filters.

\subsubsection{Fully-learned Filters and Time Conditioning}
\begin{figure*}
	\includegraphics{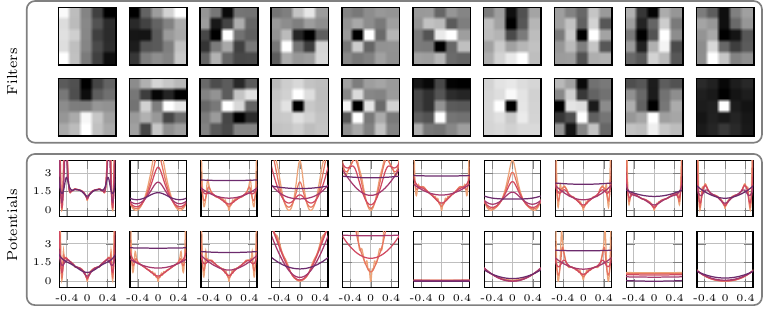}
    \caption{The learned model with Gaussian mixture factors, fully-learned zero-mean filters of size $(5 \times 5)$ and learned time conditioning.
	The variances of the Gaussian mixture factors evolve with diffusion time according to \eqref{eq:learned_embedding}. The potentials are shown for different diffusion times
	$
	\sqrt{\num{2}t} = 
	\num{0} \protect\tikz[baseline=-\the\dimexpr\fontdimen22\textfont2\relax]\protect\draw [index of colormap={0} of flare, thick] (0,0) -- (.5, 0);,
	\num{0.025} \protect\tikz[baseline=-\the\dimexpr\fontdimen22\textfont2\relax]\protect\draw [index of colormap={4} of flare, thick] (0,0) -- (.5, 0);,
	\num{0.05} \protect\tikz[baseline=-\the\dimexpr\fontdimen22\textfont2\relax]\protect\draw [index of colormap={8} of flare, thick] (0,0) -- (.5, 0);,
	\num{0.1} \protect\tikz[baseline=-\the\dimexpr\fontdimen22\textfont2\relax]\protect\draw [index of colormap={12} of flare, thick] (0,0) -- (.5, 0);,
	\num{0.2} \protect\tikz[baseline=-\the\dimexpr\fontdimen22\textfont2\relax]\protect\draw [index of colormap={17} of flare, thick] (0,0) -- (.5, 0);
	$.}
	\label{fig:learned_model_temb}
\end{figure*}

So far, we only discussed models that implement the time conditioning detailed in \eqref{eq:1dgmm_var_adapt}, which is motivated by the theory, but effectively acts as a heuristic since we cannot have ideal filters and choose \( \nu_k^2 = \norm{|\mathbf{F}\mathbf{f}_k|}_\infty \) as a practical workaround. Now, we discuss a model that learns the time conditioning as described in~\Cref{ssec:alternative_pogmdm_parametrizations}. \Cref{fig:learned_model_temb} illustrates the learned filters and their corresponding potentials. As in the previous model, several filters resemble classical ones--for instance, the fifth filter in the first row and the fourth filter in the second row resemble finite-difference filters.

\begin{figure}
	\includegraphics{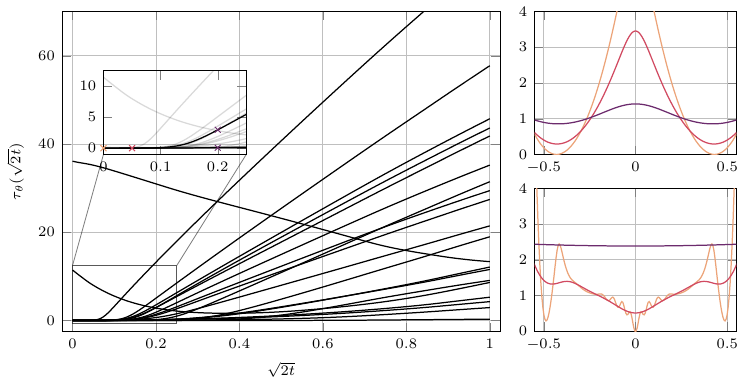}
	\caption{
	Time conditioning according to~\eqref{eq:learned_embedding} for the model with fully-learned $(5 \times 5)$ filters and time conditioning in the interval $[0, 1]$ in black (left). 
	The inset shows a detailed view of the interval $[0, 0.25]$, highlighting the time conditioning for \( \psi_{1} \) and \( \psi_{2} \). 
	The right plot shows the potentials ($-\log \psi_1$ top, $-\log \psi_{2}$ bottom) at diffusion times $\sqrt{2t} = 
	\num{0} \protect\tikz[baseline=-\the\dimexpr\fontdimen22\textfont2\relax]\protect\draw [index of colormap={0} of flare, thick] (0,0) -- (.5, 0);, 
	\num{0.05} \protect\tikz[baseline=-\the\dimexpr\fontdimen22\textfont2\relax]\protect\draw [index of colormap={8} of flare, thick] (0,0) -- (.5, 0);, 
	\num{0.2} \protect\tikz[baseline=-\the\dimexpr\fontdimen22\textfont2\relax]\protect\draw [index of colormap={17} of flare, thick] (0,0) -- (.5, 0);$.
	The corresponding values of the time conditioning are marked in the inset plot with $\times$ in the same colors.
	}
	\label{fig:output_time_embedding}
\end{figure}

Similar to the model with fully-learned convolutional filters and the original time conditioning, some potentials exhibit maxima at zero. However, most of them differ substantially from those obtained using the time conditioning in \eqref{eq:1dgmm_var_adapt}. In particular, certain potentials equilibrate rapidly as diffusion time progresses, while others evolve slowly. \Cref{fig:output_time_embedding} shows the output of the learned time-embedding network \( \tau_\theta \) for all factors over the interval \( \sqrt{2t} \in [0, 1] \). Here, we observe that most potentials begin to smooth at different diffusion times, with their time conditioning following an approximately linear progression with time.
This is especially apparent in the output for \( \psi_1 \) which increases only slightly in the interval \( [0, 0.2] \), whereas the output for \( \psi_2 \) increases rapidly after approximately \( \sqrt{2t} = 0.1 \). As a result, \( -\log \psi_2(\, \cdot\,, \mathbf{w}_k, t) \) is almost constant in the interval \( [-0.5, 0.5] \) for \( \sqrt{2t} = 0.2 \). This is highlighted in the bottom-right part of~\Cref{fig:output_time_embedding}.

\begin{figure}
	\includegraphics{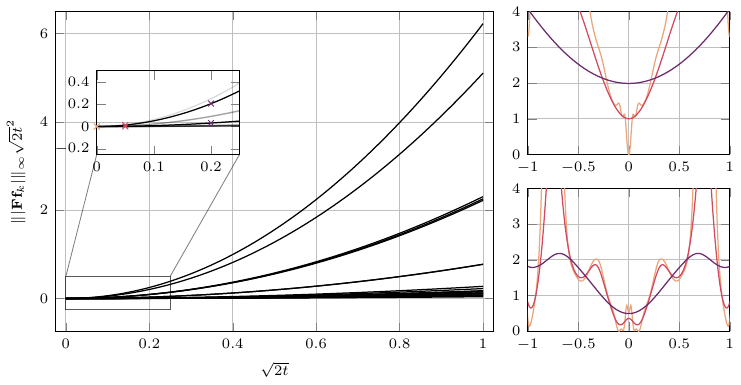}
	\caption{
	Time conditioning according to~\eqref{eq:1dgmm_var_adapt} for the convolutional model with fully-learned $(5 \times 5)$ filters in the interval $[0, 1]$ in black (left). 
	The inset shows a detailed view of the interval $[0, 0.25]$, highlighting the time conditioning for \( \psi_{15} \) and \( \psi_2 \). 
	The right plot shows the potentials ($-\log \psi_{15}$ top, $-\log \psi_{2}$ bottom) at diffusion times $\sqrt{2t} = 
	\num{0} \protect\tikz[baseline=-\the\dimexpr\fontdimen22\textfont2\relax]\protect\draw [index of colormap={0} of flare, thick] (0,0) -- (.5, 0);, 
	\num{0.05} \protect\tikz[baseline=-\the\dimexpr\fontdimen22\textfont2\relax]\protect\draw [index of colormap={8} of flare, thick] (0,0) -- (.5, 0);, 
	\num{0.2} \protect\tikz[baseline=-\the\dimexpr\fontdimen22\textfont2\relax]\protect\draw [index of colormap={17} of flare, thick] (0,0) -- (.5, 0);$. The corresponding values of the time conditioning are marked in the inset plot with $\times$ in the same colors.
	}
	\label{fig:theoretical_time_embedding}
\end{figure}
More specifically, if \( -\log \psi_k(\, \cdot\,, \mathbf{w}_k, t) = c \) for some constant \( c \in \mathbb{R}_+ \) on a given interval, then \( \psi_k(\, \cdot\,, \mathbf{w}_k, t) = \exp(-c) \) on that interval. Hence, the potential is independent of the filter response whenever it lies in this range, and contributes only a multiplicative constant to the product of the experts. Interestingly, the time-embedding network appears to learn to ignore certain filters entirely (\textit{e.g.}, the sixth and ninth filters in the second row of \Cref{fig:learned_model_temb}), as also reflected in~\Cref{fig:output_time_embedding}, where the outputs of these potentials remain substantially above zero even for \( \sqrt{2t} = 0 \).

\Cref{fig:theoretical_time_embedding} shows the adaption of the one-dimensional Gaussian mixture factor variances for the model with fully-learned filters and the original time conditioning. We immediately observe that---for the choice \(\nu_k = \norm{\mid\mathbf{F}\mathbf{f}_k\mid}_\infty \)---the time conditioning for most of the factors grows very slowly. This explains the observations in~\Cref{fig:learned_model_conv}, where many of the potentials show little variation with diffusion time. We compare two examples for a slowly and fast changing potential with diffusion time in the right part of~\Cref{fig:theoretical_time_embedding}.

Finally, we note that the learning of this model is likely suboptimal, as it does not appear to fully exploit its capacity (\textit{e.g.}, see the sixth and ninth potentials in the second row of~\Cref{fig:learned_model_temb}). Addressing this limitation is left for future work.

\subsection{Synthetic Single-Coil Data}
\label{ssec:simulation_study}
We now turn to reconstruction results and assess the performance of the models on a synthetic single-coil experiment. To this end, we simulate measurements as \( \mathbf{y} = \mathbf{M}\mathbf{F}\mathbf{x} + \mathbf{n}\), where \(\mathbf{x}\) is a normalized \gls{corpd} reference \gls{rss} reconstruction and \( \mathbf{n} \) is Gaussian noise with standard deviation \( \sigma = 0.02 \) added to the synthetic data. To account for this setup in the reconstruction algorithm, we fix the coil sensitivities to the identity and do not update them in the iterations. Furthermore, we only consider the real part of the iterates.  
We compare our approach against a classical variational penalty, an end-to-end trained baseline and a state-of-the art diffusion
based approach. As a variational baseline we consider the Charbonnier \( \varepsilon \)-smoothed isotropic \gls{tv}
\begin{equation}
	\text{TV}: \mathbf{x} \mapsto \lambda \sum_{i=1}^{d} \sqrt{(\mathbf{D}\mathbf{x})_{i}^2 + (\mathbf{D}\mathbf{x})_{i+d}^2 + \varepsilon^2}.
\end{equation}
For the end-to-end method, we choose the fastMRI baseline U-Net~\cite{knoll2020fastmri} and for the diffusion based method, we select scoreMRI~\cite{chung2022scoreMRI}. As a reference point, we additionally report results for a naive \gls{zf} reconstruction, where missing k-space data are replaced by zero prior to inverting the Fourier transformation. For both the U-Net and scoreMRI, we use the pretrained models provided by the authors. The parameters for the reverse diffusion process for scoreMRI are set to \(N=2000\), \(M=1\), \(r=0.16\), \(\sigma_\text{min}=0.01\) and \(\sigma_\text{max}=378.00\) following the choices in~\cite{chung2022scoreMRI}. Furthermore, we employ the same acceleration technique as described in~\Cref{ssec:comp_eval_methods}, and set \(t_0 = 0.2T\). The optimal regularization parameter for \gls{tv} is selected for each k-space trajectory using grid search on the validation dataset.

\Cref{tab:single_coil} shows numerical results for the single-coil experiment with synthetic data for four different k-space trajectories. ScoreMRI beats all comparison method across all k-space trajectories. Here, the gap between our best method and scoreMRI ranges from \(0.35 \) dB in the spiral case to \( 1.13 \) dB in the Cartesian case. However, the superior performance of scoreMRI comes at the cost of learning \( 6.7 \times 10^7 \) parameters during training and slow inference. Specifically, our method requires roughly two seconds per reconstruction, compared to approximately 129 seconds for scoreMRI on the same NVIDIA RTX 4090 GPU. All of our models consistently outperform \gls{tv}, except for the Gaussian k-space trajectory, where the shearlet model achieves a \( 0.03 \) dB lower \gls{psnr} compared to \gls{tv}. Remarkably, both of our models with fully-learned \((5 \times 5)\) filters beat the end-to-end U-Net baseline for a Cartesian k-space trajectory, which is setting the U-Net was trained on. \\

\setlength{\tabcolsep}{1pt}
\begin{table}[t]
    \centering
    \caption{%
        Quantitative reconstruction results for the synthetic single-coil experiment.
		The rows alternate between \gls{psnr}, \gls{ssim} and \gls{nmse}. The \gls{nmse} is scaled by $10^2$. All metrics are shown as mean \( \pm \) unit standard deviation. Bold typeface indicates the best method. Our methods: Shearlet (SH), fully-learned filters (FLF) and fully-learned filters and time conditioning (FLF+TC). Comparison methods: Zero filled (ZF), total variation (TV), U-Net~\cite{knoll2020fastmri} and scoreMRI~\cite{chung2022scoreMRI}.
    }
    \begin{tabular*}{\textwidth}{cc@{\extracolsep{\fill}}ccccccc}
        \toprule
        \multirow{2}{*}{T} & \multirow{2}{*}{A} &  \multirow{2}{*}{ZF} & \multirow{2}{*}{TV} & \multirow{2}{*}{U-Net} & \multirow{2}{*}{scoreMRI} &  \multicolumn{3}{c}{Ours}  \\
		\cmidrule(lr){7-9}
		& & & & & & SH & FLF & FLF+TC \\ 
        \midrule
        \multirow{3}{*}{C}
        & \multirow{3}{*}{4} 
		& $24.09 \pm 2.88$ & $28.87 \pm 2.16$ & $30.74 \pm 0.98$ & $\mathbf{33.29 \boldsymbol{\pm} 1.34}$ & $30.11 \pm 1.73$ & $32.08 \pm 1.52$ & $32.16 \pm 1.56$ \\
        & & $0.62 \pm 0.05$ & $0.75 \pm 0.04$ & $0.75 \pm 0.01$ & $\mathbf{0.83 \boldsymbol{\pm} 0.02}$ & $0.78 \pm 0.03$ & $0.82 \pm 0.03$ & $\mathbf{0.83 \boldsymbol{\pm} 0.03}$ \\
        & & $6.07 \pm 7.13$ & $1.65 \pm 0.99$ & $0.97 \pm 0.24$ & $\mathbf{0.54 \boldsymbol{\pm} 0.13}$ & $1.14 \pm 0.31$ & $0.72 \pm 0.20$ & $0.71 \pm 0.20$ \\
        \midrule
        \multirow{3}{*}{S}
        & \multirow{3}{*}{5} 
		& $21.42 \pm 2.39$ & $28.99 \pm 1.91$ & $26.80 \pm 1.18$ & $\mathbf{34.08 \boldsymbol{\pm} 1.22}$ & $31.65 \pm 1.38$ & $33.65 \pm 1.37$ & $33.73 \pm 1.39$ \\
        & & $0.57 \pm 0.04$ & $0.76 \pm 0.03$ & $0.68 \pm 0.02$ & $0.84 \pm 0.02$ & $0.81 \pm 0.02$ & $\mathbf{0.85 \boldsymbol{\pm} 0.02}$ & $\mathbf{0.85 \boldsymbol{\pm} 0.02}$ \\
        & & $9.85 \pm 7.13$ & $1.57 \pm 0.84$ & $2.41 \pm 0.53$ & $\mathbf{0.45 \boldsymbol{\pm} 0.11}$ & $0.79 \pm 0.17$ & $0.50 \pm 0.12$ & $0.49 \pm 0.12$ \\
        \midrule
        \multirow{3}{*}{R}
        & \multirow{3}{*}{6} 
		& $24.74 \pm 2.59$ & $30.31 \pm 1.93$ & $28.44 \pm 1.21$ & $\mathbf{33.11 \boldsymbol{\pm} 1.53}$ & $31.34 \pm 1.65$ & $32.56 \pm 1.64$ & $32.37 \pm 1.65$ \\
        & & $0.62 \pm 0.04$ & $0.78 \pm 0.03$ & $0.69 \pm 0.02$ & $\mathbf{0.82 \boldsymbol{\pm} 0.03}$ & $0.79 \pm 0.03$ & $\mathbf{0.82 \boldsymbol{\pm} 0.03}$ & $\mathbf{0.82 \boldsymbol{\pm} 0.03}$ \\
        & & $4.95 \pm 5.77$ & $1.13 \pm 0.52$ & $1.65 \pm 0.36$ & $\mathbf{0.57 \boldsymbol{\pm} 0.14}$ & $0.85 \pm 0.19$ & $0.64 \pm 0.16$ & $0.67 \pm 0.16$ \\
        \midrule
        \multirow{3}{*}{G}
        & \multirow{3}{*}{8} 
		& $21.88 \pm 1.87$ & $30.34 \pm 3.18$ & $23.68 \pm 1.90$ & $\mathbf{33.84 \boldsymbol{\pm} 1.54}$ & $33.31 \pm 1.64$ & $33.36 \pm 1.68$ & $33.36 \pm 1.74$ \\
        & & $0.59 \pm 0.05$ & $\mathbf{0.85 \boldsymbol{\pm} 0.03}$ & $0.63 \pm 0.04$ & $0.84 \pm 0.03$ & $0.84 \pm 0.03$ & $\mathbf{0.85 \boldsymbol{\pm} 0.03}$ & $\mathbf{0.85 \boldsymbol{\pm} 0.03}$ \\
        & & $7.73 \pm 2.74$ & $1.36 \pm 1.23$ & $5.04 \pm 1.56$ & $\mathbf{0.48 \boldsymbol{\pm} 0.12}$ & $0.54 \pm 0.13$ & $0.54 \pm 0.14$ & $0.54 \pm 0.13$ \\
        \midrule
        \multicolumn{3}{c}{\multirow{2}{*}{\makecell{Number of\\Parameters}}} & \multirow{2}{*}{-} & \multirow{2}{*}{$4.9 \times 10^8$} & \multirow{2}{*}{$6.7 \times 10^7$} & \multirow{2}{*}{1558} & \multirow{2}{*}{1760} & \multirow{2}{*}{7348} \\
		& & & & & & & & \\ 
        \bottomrule
    \end{tabular*}
	\footnotetext{T: k-space Trajectory, C: Cartesian, S: Spiral, R: Radial, G: 2D Gaussian, A: Acceleration}
    \label{tab:single_coil}
\end{table}
Comparing within our models, both variants with fully-learned \((5 \times 5)\) filters consistently outperform the shearlet model, achieving improvements of roughly 2 dB in the Cartesian and spiral cases.  
In the variant without learned time conditioning, this gain comes at the cost of just learning 200 additional parameters.  
The model that jointly learns the time conditioning and the convolutional filters yields a marginal improvement over the variant with the original time conditioning for the Cartesian and spiral k-space trajectories.  
The performance is identical for the Gaussian k-space trajectory, and the original time conditioning even slightly outperforms the learned version for the radial k-space trajectory.

\subsection{Parallel Imaging}%
\label{ssec:parallel_imaging}
We now evaluate our approach on real data. We compare against three baselines: (i) the Charbonnier \(\varepsilon\)-smoothed isotropic \gls{tv} embedded in the joint-nonlinear inversion algorithm of~\cite{zach2023stabledeepmri}, (ii) the end-to-end \gls{vn} of~\cite{sriram2020EndToEndVN}, which extends~\cite{hammernik2018variationalnetworks} to joint \gls{mri} reconstruction, and (iii) the end-to-end learned joint reconstruction baseline~\Gls{djs}~\cite{arvinte2021deepjsense}. We use the publicly available pretrained weights for the~\gls{vn}, retrain~\gls{djs} on the same training split as our method (see~\Cref{sec:appendix_deep_jsense_training} for further details), and determine the optimal regularization parameters for~\gls{tv} for each k-space trajectory using the procedure described in~\cite{zach2023stabledeepmri}.
The diffusion-based method scoreMRI from the synthetic single-coil experiment is not included here, as it does not support arbitrarily sized inputs due to the architecture of the score network, which is required in this experiment. Again, we report the \gls{zf} reconstruction as a reference point. Finally, to approximately map the data to intensities matching the training distribution, we normalize the measurements by \(\mathbf{y} \mapsto \mathbf{y}/\norm{\mathbf{x}_{\text{ZF}}}_\infty\) and rescale the final magnitude reconstruction as \(\mathbf{x}_0 \mapsto |\mathbf{x}_0|\norm{\mathbf{x}_{\text{ZF}}}_\infty\). Here we denote the~\gls{rss}~\gls{zf} reconstruction as \(\mathbf{x}_{\text{ZF}} = \sqrt{\sum_{i=1}^c |\mathbf{F}^H\mathbf{M}^H(\mathbf{y}_i)|^2}\).

\subsubsection{In-Distribution}
We report the quantitative results for \gls{pi} reconstruction on the \gls{corpd} dataset in \Cref{tab:pi_corpd}. Our methods consistently outperform \gls{tv} across all k-space trajectories, achieving improvements of up to \( 3.4 \) dB in the Cartesian case with rotated phase-encoding direction and the case with less available \glspl*{acl}. The \gls{vn} achieves the highest performance in its training setting (Cartesian k-space trajectory with 8\% \glspl*{acl}). In this scenario, our best method attains a \gls{psnr} 1.42 dB lower. 

\begin{figure*}
	\centering
	\includegraphics{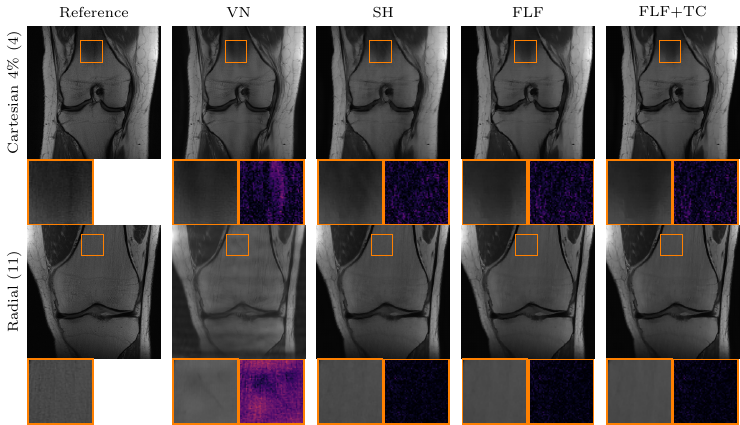}
    \caption{Qualitative \gls{pi} comparison on the \gls{corpd} dataset against the \gls{vn} in the failure case.
	First row: 4-fold Cartesian k-space trajectory with $4\%$ available \glspl*{acl} and
	an acceleration of 4. Second row: Radial k-space trajectory with an
   	acceleration of 11. Here the \gls{vn} introduces structures that are not existent in the \gls{rss} reference. 
	The inset shows a zoom on the image with corresponding absolute error. Our methods: Shearlet (SH), fully-learned filters (FLF) and fully-learned filters and time conditioning (FLF+TC).}
	\label{fig:qualitative_CORPD_comp_VN}
\end{figure*}

The performance of the \gls{vn} deteriorates significantly even under minor deviations from its training setting, for instance when the fraction of \glspl*{acl} is reduced from 8\% to 4\%. An example of this is shown in~\Cref{fig:qualitative_CORPD_comp_VN}, where the \gls{vn} struggles to produce artifact-free reconstructions, even in the Cartesian setting, and introduces artifacts in the femur and tibia. For the radial k-space trajectory, these artifacts become more severe and accompanied by hallucinations, as highlighted in the inset of~\Cref{fig:qualitative_CORPD_comp_VN}. 

\begin{figure*}
	\centering
	\includegraphics{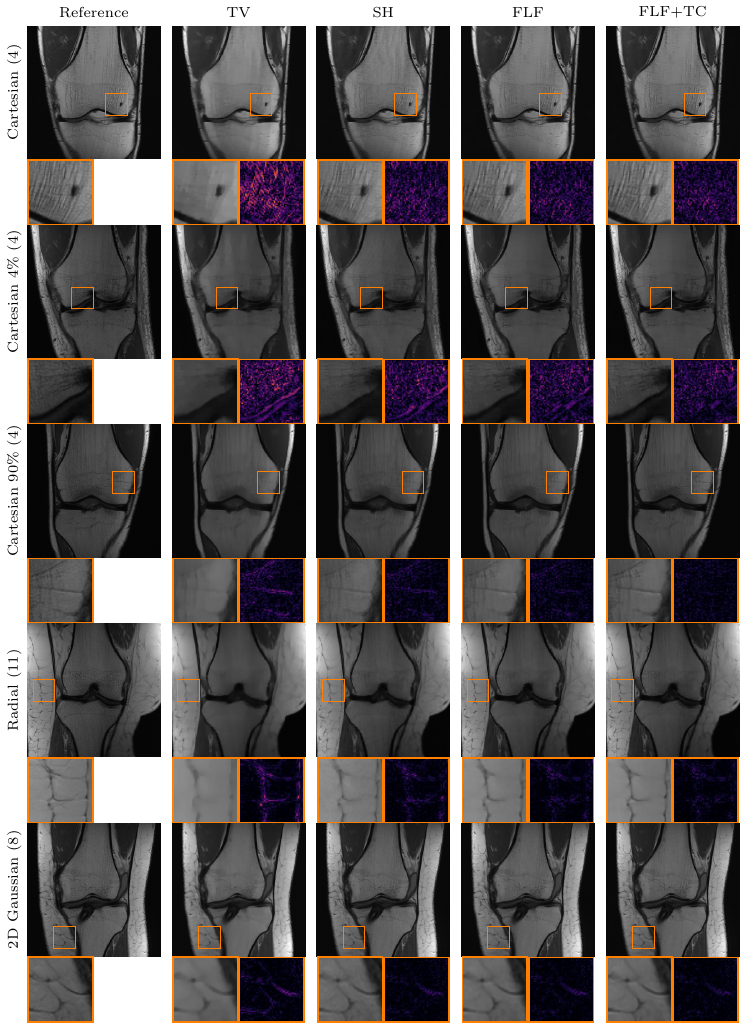}
    \caption{Qualitative \gls{pi} comparison on the \gls{corpd} dataset against \gls{tv}.
	First row: 4-fold Cartesian k-space trajectory with $8\%$ available \glspl*{acl} and
	an acceleration of 4. Second row: Same as row one with $4\%$ available \glspl*{acl}. Third row: Same as row one with rotated phase-encoding direction. Fourth row: 2D Gaussian k-space trajectory with an acceleration of 8. Fifth row: Radial k-space trajectory with an
   	acceleration of 11. The inset shows a zoom on the image with corresponding absolute error. Our methods: Shearlet (SH), fully-learned filters (FLF) and fully-learned filters and time conditioning (FLF+TC).}
	\label{fig:qualitative_CORPD}
\end{figure*}

\Cref{fig:qualitative_CORPD} shows qualitative results for the in-distribution setting against~\gls{tv}. In the prototypical case of a Cartesian k-space trajectory with \(8 \%\) available \glspl*{acl}, the reconstruction obtained with \gls{tv} appears largely artifact-free, but the method fails to recover fine details. In contrast, the reconstructions obtained with the shearlet model show more small scale details which are also visible in the \gls{rss} reference. However, in some reconstructions slight backfolding artifacts, such as those visible in the upper part of the femur, remain. Both models which employ fully-learned convolutional filters produce artifact-free reconstructions, and the methods retain even finer details such as the sharper vertical edge visible in the zoomed regions of the two rightmost reconstructions in~\Cref{fig:qualitative_CORPD}. With a reduced number of \glspl*{acl}, the \gls{tv} reconstruction exhibits pronounced backfolding artifacts in the femur, again our models remain largely robust.
Furthermore, for the Cartesian with rotated phase-encoding and radial k-space trajectory, the models with fully-learned filters produce less noisy reconstructions and enhance structural details such as vessels in the fat tissue better. For the Gaussian k-space trajectory, all of our methods yield comparable results. This is likely due to the fact that the measured data inherently retains substantial amounts of low- and high-frequency information (as evident by the zero-filled solution which already achieves \( 32.15 \) dB in~\Cref{tab:pi_corpd}).

\begin{figure*}
	\centering
	\includegraphics{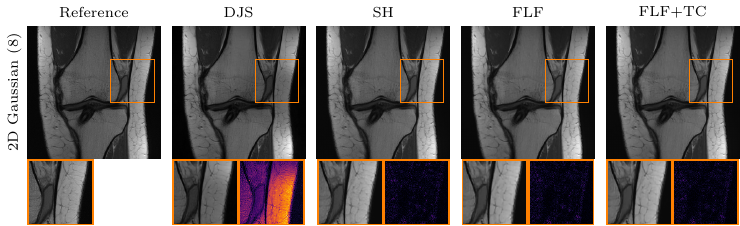}
    \caption{Qualitative \gls{pi} comparison on the \gls{corpd} dataset against \gls{djs} for a two-dimensional Gaussian k-space trajectory with and acceleration of 8.
	\Gls{djs} introduces contrast artifacts in the lateral fat tissue, due to suboptimal coil sensitivity estimation. In contrast, these are not present in the reconstructions of our models.
	The inset shows a zoom on the image with corresponding absolute error. Our methods: Shearlet (SH), fully-learned filters (FLF) and fully-learned filters and time conditioning (FLF+TC).}
	\label{fig:qualitative_CORPD_comp_DJS}
\end{figure*}

\Gls{djs} outperforms our method in terms of~\gls{ssim} in both regular Cartesian settings. However, its mean~\gls{psnr} and~\gls{nmse} are inferior. Interestingly,~\gls{djs} demonstrates greater robustness to changes in the forward model compared to the~\gls{vn}. This is both reflected in the radial- and the Cartesian k-space trajectory with rotated phase-encoding, where~\gls{djs} significantly outperforms the~\gls{vn}, and achieves a~\gls{ssim} comparable to our method. For the Gaussian k-space trajectory, \gls{djs} produces plausible reconstructions but struggles with accurate coil sensitivity estimation. This limitation introduces noticeable artifacts, examples of which are shown in~\Cref{fig:qualitative_CORPD_comp_DJS}. In contrast, our methods maintain robustness across all variations.

\begin{table}[t]
\centering
\footnotesize
\caption{Quantitative \gls{pi} reconstruction results on the in-distribution knee dataset with proton density contrast.
The rows alternate between \gls{psnr}, \gls{ssim} and \gls{nmse}. The \gls{nmse} is scaled by $10^2$. All metrics are shown as mean $\pm$ unit standard deviation. 
Bold typeface indicates the best method. Our methods: Shearlet (SH), fully-learned filters (FLF) and fully-learned filters and time conditioning (FLF+TC). Comparison methods: Zero filled (ZF), total variation (TV)~\cite{zach2023stabledeepmri}, end-to-end variational network (VN)~\cite{sriram2020EndToEndVN} and Deep-JSense (DJS)~\cite{arvinte2021deepjsense}.}
\begin{tabular*}{\textwidth}{@{\extracolsep\fill}cccccccccc}
\toprule
\multirow{2}{*}{T} & \multirow{2}{*}{A} & \multirow{2}{*}{ACL} 
& \multirow{2}{*}{ZF} & \multirow{2}{*}{TV} & \multirow{2}{*}{VN} & \multirow{2}{*}{DJS} & \multicolumn{3}{c}{Ours} \\
\cmidrule(lr){8-10}
& & & & & & & SH & FLF & FLF+TC \\
\midrule
\multirow{9}{*}{C}
& \multirow{9}{*}{4} & \multirow{3}{*}{8} 
& $27.19 \pm 1.85$ & $33.00 \pm 1.92$ & $\mathbf{36.92 \boldsymbol{\pm} 2.21}$ & $34.45 \pm 2.48$ & $34.64 \pm 1.96$ & $35.39 \pm 2.20$ & $35.50 \pm 2.65$ \\
& & & $0.74 \pm 0.04$ & $0.83 \pm 0.04$ & $\mathbf{0.92 \boldsymbol{\pm} 0.03}$ & $0.90 \pm 0.04$ & $0.88 \pm 0.03$ & $0.89 \pm 0.03$ & $0.89 \pm 0.04$ \\
& & & $2.37 \pm 0.72$ & $0.59 \pm 0.15$ & $\mathbf{0.24 \boldsymbol{\pm} 0.09}$ & $0.46 \pm 0.27$ & $0.41 \pm 0.13$ & $0.36 \pm 0.20$ & $0.36 \pm 0.32$ \\
\cmidrule(l){3-10}
& & \multirow{3}{*}{8\footnotemark[1]} 
& $31.13 \pm 2.36$ & $33.65 \pm 2.09$ & $24.72 \pm 1.90$ & $33.20 \pm 5.71$ & $36.19 \pm 2.18$ & $36.68 \pm 1.95$ & $\mathbf{37.05 \boldsymbol{\pm} 2.20}$ \\
& & & $0.81 \pm 0.05$ & $0.84 \pm 0.05$ & $0.67 \pm 0.06$ & $0.90 \pm 0.08$ & $0.91 \pm 0.03$ & $0.91 \pm 0.03$ & $\mathbf{0.92 \boldsymbol{\pm} 0.03}$ \\
& & & $0.97 \pm 0.40$ & $0.51 \pm 0.15$ & $3.29 \pm 1.00$ & $1.85 \pm 6.84$ & $0.29 \pm 0.11$ & $0.26 \pm 0.09$ & $\mathbf{0.24 \boldsymbol{\pm} 0.10}$ \\
\cmidrule(l){3-10}
& & \multirow{3}{*}{4} 
& $24.14 \pm 1.66$ & $31.86 \pm 2.41$ & $32.16 \pm 1.62$ & $33.84 \pm 2.44$ & $34.30 \pm 1.77$ & $35.29 \pm 2.14$ & $\mathbf{35.32 \boldsymbol{\pm} 2.53}$ \\
& & & $0.69 \pm 0.04$ & $0.82 \pm 0.05$ & $0.89 \pm 0.03$ & $\mathbf{0.90} \boldsymbol{\pm} \mathbf{0.04}$ &  $0.88 \pm 0.03$ & $0.89 \pm 0.04$ & $0.89 \pm 0.04$ \\
& & & $4.96 \pm 1.52$ & $0.81 \pm 0.35$ & $0.69 \pm 0.15$ & $0.53 \pm 0.29$ &  $0.44 \pm 0.12$ & $\mathbf{0.36 \boldsymbol{\pm} 0.17}$ & $\mathbf{0.36 \boldsymbol{\pm} 0.28}$ \\
\midrule
\multirow{3}{*}{R}
& \multirow{3}{*}{11} & \multirow{3}{*}{-} 
& $28.76 \pm 2.15$ & $33.24 \pm 1.92$ & $20.56 \pm 1.72$ & $32.79 \pm 2.69$ & $34.54 \pm 2.04$ & $34.94 \pm 2.03$ & $\mathbf{35.10 \boldsymbol{\pm} 2.28}$ \\
& & & $0.75 \pm 0.06$ & $0.82 \pm 0.04$ & $0.69 \pm 0.05$ & $0.86 \pm 0.04$ & $\mathbf{0.87 \boldsymbol{\pm} 0.04}$ & $\mathbf{0.87 \boldsymbol{\pm} 0.04}$ & $\mathbf{0.87 \boldsymbol{\pm} 0.04}$ \\
& & & $1.67 \pm 0.58$ & $0.56 \pm 0.16$ & $6.55 \pm 1.14$ & $0.71 \pm 0.55$ & $0.42 \pm 0.14$ & $\mathbf{0.38 \boldsymbol{\pm} 0.13}$ & $\mathbf{0.38 \boldsymbol{\pm} 0.16}$ \\
\midrule
\multirow{3}{*}{G}
& \multirow{3}{*}{8} & \multirow{3}{*}{-} 
& $32.15 \pm 2.32$ & $34.35 \pm 1.99$ & $23.53 \pm 1.94$ & $25.44 \pm 3.69$ & $35.60 \pm 2.20$ & $35.75 \pm 2.06$ & $\mathbf{35.84 \boldsymbol{\pm} 2.17}$ \\
& & & $0.84 \pm 0.05$ & $0.86 \pm 0.04$ & $0.73 \pm 0.05$ & $0.80 \pm 0.06$ & $\mathbf{0.89 \boldsymbol{\pm} 0.04}$ & $\mathbf{0.89 \boldsymbol{\pm} 0.03}$ & $\mathbf{0.89 \boldsymbol{\pm} 0.03}$ \\
& & & $0.78 \pm 0.35$ & $0.44 \pm 0.12$ & $4.33 \pm 1.19$ & $6.63 \pm 6.69$ & $0.33 \pm 0.12$ & $\mathbf{0.32 \boldsymbol{\pm} 0.11}$ & $\mathbf{0.32 \boldsymbol{\pm} 0.11}$ \\
\midrule
\multicolumn{3}{c}{\multirow{2}{*}{\footnotesize \makecell{Number of\\Parameters}}} 
& \multirow{2}{*}{-} & \multirow{2}{*}{-} & \multirow{2}{*}{$3\times 10^{7}$} & \multirow{2}{*}{$447490$} & \multirow{2}{*}{$1558$} & \multirow{2}{*}{$1760$} & \multirow{2}{*}{$7348$} \\
& & & & & & & & & \\
\bottomrule
\end{tabular*}
\footnotetext{\textsuperscript{a} Rotated phase-encoding direction.}
\footnotetext{T: k-space Trajectory, C: Cartesian, R: Radial, G: 2D Gaussian, A: Acceleration, ACL: Auto Calibration Lines}
\label{tab:pi_corpd}
\end{table}

Comparing the shearlet model with the two models with fully-learned convolutional filters, we observe that the former is inferior for all considered k-space trajectories. Specifically, the model with additionally learned time conditioning achieves an average improvement of \(0.7\) dB over the shearlet model, while outperforming the model with fully-learned convolutional filters and the original time conditioning by \(0.15\) dB on average.

\subsubsection{Contrast Out-of-Distribution} 
\begin{table}[t]
\centering
\footnotesize
\caption{Quantitative results for PI experiments on the contrast out-of-distribution dataset (CORPDFS).
The rows alternate between \gls{psnr}, \gls{ssim} and \gls{nmse}. The \gls{nmse} is scaled by $10^2$. All metrics are shown as mean $\pm$ unit standard deviation. 
Bold typeface indicates the best method. Our methods: Shearlet (SH), fully-learned filters (FLF) and fully-learned filters and time conditioning (FLF+TC). Comparison methods: Zero filled (ZF), total variation (TV)~\cite{zach2023stabledeepmri}, end-to-end variational network (VN)~\cite{sriram2020EndToEndVN} and Deep-JSense (DJS)~\cite{arvinte2021deepjsense}.}
\begin{tabular*}{\textwidth}{@{\extracolsep\fill}cccccccccc}
\toprule
\multirow{2}{*}{T} & \multirow{2}{*}{A} & \multirow{2}{*}{ACL} 
& \multirow{2}{*}{ZF} & \multirow{2}{*}{TV} & \multirow{2}{*}{VN} & \multirow{2}{*}{DJS} & \multicolumn{3}{c}{Ours} \\
\cmidrule(lr){8-10}
& & & & & & & SH & FLF & FLF+TC \\
\midrule
\multirow{9}{*}{C}
& \multirow{9}{*}{4} & \multirow{3}{*}{8} 
& $26.32 \pm 2.75$ & $31.47 \pm 2.58$ & $30.28 \pm 3.67$ & $11.31 \pm 12.95$ & $31.78 \pm 2.65$ & $\mathbf{32.06 \boldsymbol{\pm} 2.78}$ & $31.50 \pm 3.10$ \\
& & & $0.68 \pm 0.08$ & $0.74 \pm 0.10$ & $\mathbf{0.77 \boldsymbol{\pm} 0.09}$ & $0.33 \pm 0.26$ & $0.76 \pm 0.09$ & $\mathbf{0.77 \boldsymbol{\pm} 0.10}$ & $0.75 \pm 0.11$\\
& & & $7.08 \pm 4.07$ & $1.66 \pm 0.97$ & $2.90 \pm 2.65$ & $5.2e4 \pm 2.8e5$ & $1.60 \pm 1.54$ & $\mathbf{1.52 \boldsymbol{\pm} 1.53}$ & $1.72 \pm 1.54 $\\
\cmidrule(l){3-10}
& & \multirow{3}{*}{8\footnotemark[1]} 
& $26.68 \pm 3.14$ & $31.97 \pm 2.70$ & $28.73 \pm 2.70$ & $11.07 \pm 11.09$ & $32.26 \pm 2.77$ & $\mathbf{32.46 \boldsymbol{\pm} 2.75}$ & $32.27 \pm 2.94$\\
& & & $0.71 \pm 0.08$ & $0.75 \pm 0.10$ & $0.71 \pm 0.08$ & $0.21 \pm 0.22$ & $\mathbf{0.78 \boldsymbol{\pm} 0.09}$ & $0.77 \pm 0.09$ & $0.77 \pm 0.10$\\
& & & $7.03 \pm 4.95$ & $1.51 \pm 1.08$ & $3.15 \pm 1.97$ & $5.8e6 \pm 1.7e7$ & $1.46 \pm 1.54$ &  $\mathbf{1.40 \boldsymbol{\pm} 1.54}$ & $1.47 \pm 1.52$\\
\cmidrule(l){3-10}
& & \multirow{3}{*}{4} 
& $25.19 \pm 2.33$ & $31.24 \pm 2.63$ & $29.46 \pm 3.23$ & $9.38 \pm 16.45$ & $31.65 \pm 2.58$ & $\mathbf{31.86 \boldsymbol{\pm} 2.82}$ & $31.06 \pm 3.04$\\
& & & $0.66 \pm 0.08$ & $0.73 \pm 0.10$ & $\mathbf{0.76 \boldsymbol{\pm} 0.09}$ & $0.34 \pm 0.26$ & $\mathbf{0.76 \boldsymbol{\pm} 0.09}$ & $\mathbf{0.76 \boldsymbol{\pm} 0.10}$ & $0.74 \pm 0.12$\\
& & & $9.05 \pm 4.23$ & $1.77 \pm 1.07$ & $3.32 \pm 2.76$ & $1.6e4 \pm 8.9e4$ & $1.65 \pm 1.55$ & $\mathbf{1.58 \boldsymbol{\pm} 1.52}$ & $1.86 \pm 1.48$\\
\midrule
\multirow{3}{*}{R}
& \multirow{3}{*}{11} & \multirow{3}{*}{-} 
& $25.25 \pm 3.25$ & $31.56 \pm 2.59$ & $26.41 \pm 2.14$ & $12.40 \pm 13.39$ & $31.54 \pm 2.68$ & $\mathbf{31.85 \boldsymbol{\pm} 2.72}$ & $31.26 \pm 2.89$\\
& & & $0.62 \pm 0.10$ & $0.73 \pm 0.10$ & $0.69 \pm 0.08$ & $0.33 \pm 0.25$ & $0.74 \pm 0.10$ & $\mathbf{0.75 \boldsymbol{\pm} 0.10}$ & $0.73 \pm 0.12$\\
& & & $10.94 \pm 8.48$ & $1.65 \pm 1.32$ & $3.95 \pm 1.54$ & $4.3e5 \pm 3.0e6$ & $1.68 \pm 1.56$ & $\mathbf{1.58 \boldsymbol{\pm} 1.55}$ & $1.80 \pm 1.54$\\
\midrule
\multirow{3}{*}{G}
& \multirow{3}{*}{8} & \multirow{3}{*}{-} 
& $26.86 \pm 3.84$ & $32.07 \pm 2.55$ & $28.22 \pm 2.92$ & $21.86 \pm 7.80$ & $32.09 \pm 2.68$ & $\mathbf{32.30 \boldsymbol{\pm} 2.71}$ & $32.15 \pm 2.73$\\
& & & $0.70 \pm 0.10$ & $0.75 \pm 0.10$ & $0.72 \pm 0.09$ & $0.60 \pm 0.20$ & $0.76 \pm 0.09$ & $\mathbf{0.77 \boldsymbol{\pm} 0.09}$ & $0.76 \pm 0.10$\\
& & & $7.98 \pm 6.89$ & $1.46 \pm 1.10$ & $3.76 \pm 2.84$ & $9.9e2 \pm 7.4e3$ & $1.52 \pm 1.56$ & $\mathbf{1.45 \boldsymbol{\pm} 1.55}$ & $1.50 \pm 1.54$\\
\midrule
\multicolumn{3}{c}{\multirow{2}{*}{\footnotesize \makecell{Number of\\Parameters}}} 
& \multirow{2}{*}{-} & \multirow{2}{*}{-} & \multirow{2}{*}{$3\times 10^{7}$} & \multirow{2}{*}{$447490$} & \multirow{2}{*}{$1558$} & \multirow{2}{*}{$1760$} & \multirow{2}{*}{$7348$} \\
& & & & & & & & & \\
\bottomrule
\end{tabular*}
\footnotetext{\textsuperscript{a} Rotated phase-encoding direction.}
\footnotetext{T: k-space Trajectory, C: Cartesian, R: Radial, G: 2D Gaussian, A: Acceleration, ACL: Auto Calibration Lines}
\label{tab:pi_corpdfs}
\end{table}
We now investigate the robustness of the methods to shifts in the contrast mechanism, transitioning from \gls{corpd}-weighted to \gls{corpdfs}-weighted data. We show in \Cref{tab:pi_corpdfs} the quantitative numerical results across different k-space trajectories. Interestingly, due to the shift in the contrast mechanism, the performance of the~\gls{vn} deteriorates even for the vertical Cartesian case with \(8\%\)\glspl{acl}. Similar to the results in the previous section, the performance of the \gls{vn} further decreases for all other k-space trajectories. \Gls{djs} shows no robustness to the contrast-mechanism shift, with its average performance falling below the naive zero-filled baseline. Furthermore, the high standard deviation across all metrics indicates significant reconstruction variability, \textit{i.e.} some reconstructions remain plausible, while others are contaminated by severe artifacts. This behaviour, along with the high \gls{nmse} values, are further investigated in \Cref{sec:appendix_deep_jsense_training}. In contrast, \gls{tv} remains robust and delivers good results across all cases. Among our models, the model which implements fully-learned filters performs best across all k-space trajectories and comparison methods, achieving an average improvement of around \(0.45\) dB in terms of \gls{psnr} compared to \gls{tv}. 

\begin{figure*}
	\centering
	\includegraphics{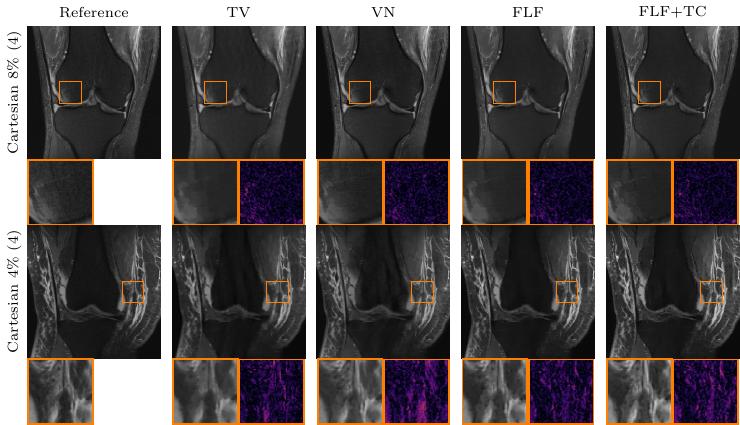}
    \caption{Qualitative \gls{pi} comparison on the \gls{corpdfs} dataset against the \gls{vn} and \gls{tv}.
	First row: 4-fold Cartesian k-space trajectory with $8\%$ available \glspl*{acl} and an acceleration of 4.
	\glspl*{acl}. Second row: 4-fold Cartesian k-space trajectory with $4\%$ \glspl*{acl} and
	an acceleration of 4. 
	The inset shows a zoom on the image with corresponding absolute error.
	Our methods: Fully-learned filters (FLF) and fully-learned filters and time conditioning (FLF+TC).}
	\label{fig:qualitative_CORPDFS_comp_VN}
\end{figure*}
Interestingly, the shearlet model and the model with fully-learned filters and time conditioning achieve performance comparable to \gls{tv}. In the Cartesian case with only \(4\%\) of available~\glspl{acl}, however, the latter shows a significant drop. We believe this and the overall performance drop for the fat-suppressed data is due to the increased noise in the data, thus impairing quantitative comparison. This interpretation is supported by the qualitative results in~\Cref{fig:qualitative_CORPDFS_comp_VN}, where the reference reconstructions appear very noisy. In contrast to~\gls{tv}, our models produce cleaner reconstructions with well-defined edges, as highlighted in the inset for the model with fully-learned filters and time conditioning. Moreover, \gls{tv} reconstructions occasionally retain undersampling artifacts, while the \gls{vn} exhibits the expected strong artifacts when the number of~\glspl{acl} is reduced.

\subsubsection{Anatomical Out-of-Distribution} 
\Cref{tab:pi_braint1} and~\Cref{tab:pi_braint2} summarize quantitative results for the out-of-distribution experiments involving shifts in both the contrast mechanism and the underlying anatomy. 
For T1-weighted brain images, our models with fully-learned convolutional filters clearly outperform all other methods, including the shearlet model. 
Specifically, our model with original time conditioning improves average \gls{psnr} by around 2 dB over \gls{tv} across all k-space trajectories. Incorporating learned time conditioning provides an additional 0.1 dB gain.

Consistent with the previous out-of-distribution experiment, the end-to-end \gls{vn} and~\gls{djs} perform worse than our methods for all k-space trajectories. Notably, \gls{djs} exhibits extreme failure in most cases, yielding negative average \gls{psnr} values and extremely high standard deviations for all but the Gaussian k-space trajectory. Again, we investigate this behaviour further in \Cref{sec:appendix_deep_jsense_training}.

\begin{figure*}
	\centering
	\includegraphics{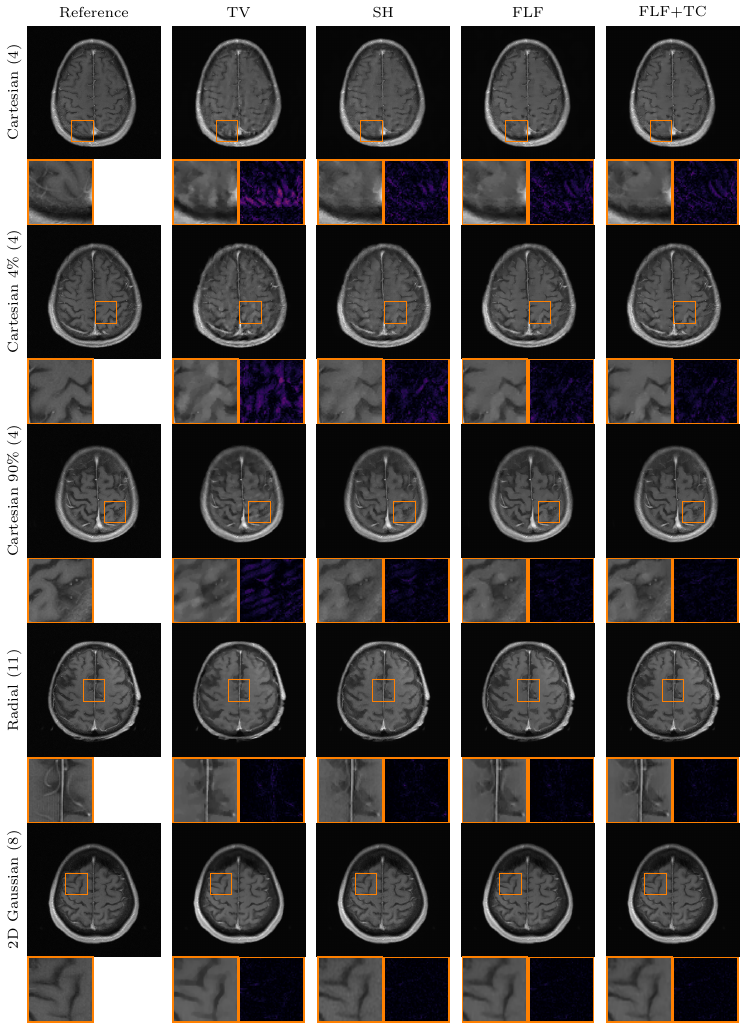}
    \caption{Qualitative \gls{pi} comparison on the brain T1 dataset against \gls{tv}.
	First row: 4-fold Cartesian k-space trajectory with $8\%$ available \glspl*{acl} and an acceleration of 4. Second row: Same as row one with $4\%$ available \glspl*{acl}. Third row: Same as row one with rotated phase-encoding direction. Fourth row: 2D Gaussian k-space trajectory with an acceleration of 8. Fifth row: Radial k-space trajectory with
   	an acceleration of 11. The inset shows a zoom on the image with corresponding absolute error. Our methods: Shearlet (SH), fully-learned filters (FLF) and fully-learned filters and time conditioning (FLF+TC).}
	\label{fig:qualitative_brain_T1}
\end{figure*}

The quantitative improvements of our method are mirrored in the qualitative reconstructions shown in~\Cref{fig:qualitative_brain_T1}. 
All reconstructions obtained with \gls{tv} and a Cartesian k-space trajectory exhibit backfolding artifacts which are clearly visible in the insets of the first and second rows of~\Cref{fig:qualitative_brain_T1}. 
In contrast, our methods resolve these artifacts even under strong distribution shifts. 
As expected, small structures such as blood vessels (\textit{e.g.}, in the inset of the radial k-space trajectory in~\Cref{fig:qualitative_brain_T1}) are not well recovered by \gls{tv}. Furthermore, both models with fully-learned convolutional filters outperform the shearlet model and produce reconstructions with more pronounced vessels and reduced noise. These observations are further supported by the one-step empirical-Bayes denoising results in~\Cref{sec:appendix_B}.

\begin{table}[t]
\centering
\footnotesize
\caption{Quantitative \gls{pi} reconstruction results on the anatomical out-of-distribution dataset with T1 contrast.
The rows alternate between \gls{psnr}, \gls{ssim} and \gls{nmse}. The \gls{nmse} is scaled by $10^2$. All metrics are shown as mean $\pm$ unit standard deviation. 
Bold typeface indicates the best method. Our methods: Shearlet (SH), fully-learned filters (FLF) and fully-learned filters and time conditioning (FLF+TC). Comparison methods: Zero filled (ZF), total variation (TV)~\cite{zach2023stabledeepmri}, end-to-end variational network (VN)~\cite{sriram2020EndToEndVN} and Deep-JSense (DJS)~\cite{arvinte2021deepjsense}.}
\begin{tabular*}{\textwidth}{@{\extracolsep\fill}cccccccccc}
\toprule
\multirow{2}{*}{T} & \multirow{2}{*}{A} & \multirow{2}{*}{ACL} 
& \multirow{2}{*}{ZF} & \multirow{2}{*}{TV} & \multirow{2}{*}{VN} & \multirow{2}{*}{DJS} & \multicolumn{3}{c}{Ours} \\
\cmidrule(lr){8-10}
& & & & & & & SH & FLF & FLF+TC \\
\midrule
\multirow{9}{*}{C}
& \multirow{9}{*}{4} & \multirow{3}{*}{8} 
& $28.49 \pm 1.74$ & $33.43 \pm 1.75$ & $32.09 \pm 1.43$ & $-8.81 \pm 40.00$ & $34.55 \pm 1.61$ & $35.65 \pm 1.54$ & $\mathbf{35.85 \boldsymbol{\pm} 1.53}$ \\
& & & $0.78 \pm 0.03$ & $0.91 \pm 0.02$ & $0.84 \pm 0.02$ & $0.41 \pm 0.33$ & $0.91 \pm 0.02$ & $0.92 \pm 0.02$ & $\mathbf{0.93 \boldsymbol{\pm} 0.02}$ \\
& & & $3.91 \pm 1.16$ & $1.19 \pm 0.36$ & $1.50 \pm 0.28$ & $55.55 \pm 38.45$ & $0.92 \pm 0.25$ & $0.71 \pm 0.20$ & $\mathbf{0.68 \boldsymbol{\pm} 0.23}$\\
\cmidrule(l){3-10}
& & \multirow{3}{*}{8\footnotemark[1]} 
& $29.37 \pm 1.71$ & $35.88 \pm 1.64$ & $20.72 \pm 1.71$ & $-48.59 \pm 26.23$ & $37.49 \pm 1.56$ & $38.28 \pm 1.51$ & $\mathbf{38.33 \boldsymbol{\pm} 1.31}$\\
& & & $0.78 \pm 0.02$ & $0.92 \pm 0.02$ & $0.44 \pm 0.04$ & $0.04 \pm 0.09$ & $\mathbf{0.94 \boldsymbol{\pm} 0.01}$ & $\mathbf{0.94 \boldsymbol{\pm} 0.01}$ & $\mathbf{0.94 \boldsymbol{\pm} 0.01}$\\
& & & $3.19 \pm 0.93$ & $0.67 \pm 0.18$ & $14.59 \pm 3.23$ & $98.19 \pm 6.36$ & $0.45 \pm 0.10$ & $\mathbf{0.38 \boldsymbol{\pm} 0.08}$ & $\mathbf{0.38 \boldsymbol{\pm} 0.12}$\\
\cmidrule(l){3-10}
& & \multirow{3}{*}{4} 
& $25.19 \pm 1.57$ & $32.01 \pm 1.71$ & $29.55 \pm 1.47$ & $-4.18 \pm 39.55$ & $33.41 \pm 1.58$ & $34.82 \pm 1.55$ & $\mathbf{35.28 \boldsymbol{\pm} 1.51}$\\
& & & $0.70 \pm 0.03$ & $0.90 \pm 0.02$ & $0.81 \pm 0.02$ & $0.45 \pm 0.32$ & $0.91 \pm 0.02$ & $0.92 \pm 0.02$ & $\mathbf{0.93 \boldsymbol{\pm} 0.02}$\\
& & & $8.73 \pm 1.85$ & $1.66 \pm 0.46$ & $2.68 \pm 0.49$ & $47.90 \pm 37.52$ & $1.21 \pm 0.31$ & $0.86 \pm 0.23$ & $\mathbf{0.78 \boldsymbol{\pm} 0.25}$\\
\midrule
\multirow{3}{*}{R}
& \multirow{3}{*}{11} & \multirow{3}{*}{-} 
& $28.50 \pm 1.52$ & $35.49 \pm 1.50$ & $16.19 \pm 1.89$ & $-46.05 \pm 28.79$ & $35.96 \pm 1.50$ & $\mathbf{36.82 \boldsymbol{\pm} 1.38}$ & $36.75 \pm 1.22 $\\
& & & $0.80 \pm 0.02$ & $0.92 \pm 0.01$ & $0.42 \pm 0.04$ & $0.05 \pm 0.15$ &$\mathbf{0.93 \boldsymbol{\pm} 0.01}$ & $\mathbf{0.93 \boldsymbol{\pm} 0.01}$ & $\mathbf{0.93 \boldsymbol{\pm} 0.01}$\\
& & & $4.01 \pm 0.96$ & $0.72 \pm 0.15$ & $27.61 \pm 7.62$ & $94.91 \pm 16.46$ & $0.64 \pm 0.12$ & $\mathbf{0.53 \boldsymbol{\pm} 0.10}$ & $0.54 \pm 0.16$\\
\midrule
\multirow{3}{*}{G}
& \multirow{3}{*}{8} & \multirow{3}{*}{-} 
& $34.86 \pm 1.44$ & $36.63 \pm 1.44$ & $22.15 \pm 1.98$ & $21.29 \pm 3.35$ & $37.29 \pm 1.52$ & $\mathbf{37.72 \boldsymbol{\pm} 1.46}$ & $37.63 \pm 1.49$\\
& & & $0.90 \pm 0.02$ & $\mathbf{0.94 \boldsymbol{\pm} 0.01}$ & $0.56 \pm 0.05$ & $0.73 \pm 0.07$ & $\mathbf{0.94 \boldsymbol{\pm} 0.01}$ & $\mathbf{0.94 \boldsymbol{\pm} 0.01}$ & $\mathbf{0.94 \boldsymbol{\pm} 0.01}$\\
& & & $0.86 \pm 0.13$ & $0.55 \pm 0.09$ & $11.86 \pm 2.50$ & $81.67 \pm 87.22$ & $0.47 \pm 0.08$ & $\mathbf{0.43 \boldsymbol{\pm} 0.07}$ & $0.44 \pm 0.08$\\
\midrule
\multicolumn{3}{c}{\multirow{2}{*}{\footnotesize \makecell{Number of\\Parameters}}} 
& \multirow{2}{*}{-} & \multirow{2}{*}{-} & \multirow{2}{*}{$3\times 10^{7}$} & \multirow{2}{*}{$447490$} & \multirow{2}{*}{$1558$} & \multirow{2}{*}{$1760$} & \multirow{2}{*}{$7348$} \\
& & & & & & & & & \\
\bottomrule
\end{tabular*}
\footnotetext{\textsuperscript{a} Rotated phase-encoding direction.}
\footnotetext{T: k-space Trajectory, C: Cartesian, R: Radial, G: 2D Gaussian, A: Acceleration, ACL: Auto Calibration Lines}
\label{tab:pi_braint1}
\end{table}

\begin{table}[t]
\centering
\footnotesize
\caption{Quantitative \gls{pi} reconstruction results on the anatomical out-of-distribution dataset with T2 contrast.
The rows alternate between \gls{psnr}, \gls{ssim} and \gls{nmse}. The \gls{nmse} is scaled by $10^2$. All metrics are shown as mean $\pm$ unit standard deviation. 
Bold typeface indicates the best method. Our methods: Shearlet (SH), fully-learned filters (FLF) and fully-learned filters and time conditioning (FLF+TC). Comparison methods: Zero filled (ZF), total variation (TV)~\cite{zach2023stabledeepmri}, end-to-end variational network (VN)~\cite{sriram2020EndToEndVN} and Deep-JSense (DJS)~\cite{arvinte2021deepjsense}.}
\begin{tabular*}{\textwidth}{@{\extracolsep\fill}cccccccccc}
\toprule
\multirow{2}{*}{T} & \multirow{2}{*}{A} & \multirow{2}{*}{ACL} 
& \multirow{2}{*}{ZF} & \multirow{2}{*}{TV} & \multirow{2}{*}{VN} & \multirow{2}{*}{DJS} & \multicolumn{3}{c}{Ours} \\
\cmidrule(lr){8-10}
& & & & & & & SH & FLF & FLF+TC \\
\midrule
\multirow{9}{*}{C}
& \multirow{9}{*}{4} & \multirow{3}{*}{8}
& $26.30 \pm 1.10$ & $32.61 \pm 1.51$ & $32.28 \pm 1.12$ & $-13.72 \pm 27.96$ & $32.09 \pm 1.27$ & $33.63 \pm 1.29$ & $\mathbf{33.81 \boldsymbol{\pm} 1.41}$ \\
& & & $0.74 \pm 0.02$ & $0.90 \pm 0.02$ & $0.88 \pm 0.02$ & $0.20 \pm 0.27$ & $0.90 \pm 0.02$ & $\mathbf{0.91 \boldsymbol{\pm} 0.02}$ & $\mathbf{0.91 \boldsymbol{\pm} 0.02}$ \\
& & & $5.59 \pm 0.91$ & $1.26 \pm 0.36$ & $1.27 \pm 0.23$ & $77.11 \pm 34.07$ & $1.41 \pm 0.27$ & $0.99 \pm 0.23$ & $\mathbf{0.94 \boldsymbol{\pm} 0.24}$\\
\cmidrule(l){3-10}
& & \multirow{3}{*}{8\footnotemark[1]} 
& $26.53 \pm 0.97$ & $35.01 \pm 1.64$ & $20.61 \pm 1.64$ & $-25.36 \pm 36.16$ & $34.55 \pm 1.39$ & $35.73 \pm 1.39$ & $\mathbf{35.92 \boldsymbol{\pm} 1.57}$\\
& & & $0.74 \pm 0.02$ & $0.92 \pm 0.02$ & $0.49 \pm 0.05$ & $0.17 \pm 0.25$ & $0.92 \pm 0.02$ & $\mathbf{0.93 \boldsymbol{\pm} 0.02}$ & $\mathbf{0.93 \boldsymbol{\pm} 0.02}$\\
& & & $5.30 \pm 0.75$ & $0.72 \pm 0.21$ & $13.35 \pm 3.07$ & $283.86 \pm 992.41$ & $0.79 \pm 0.16$ & $0.61 \pm 0.15$ & $\mathbf{0.58 \boldsymbol{\pm} 0.17}$\\
\cmidrule(l){3-10}
& & \multirow{3}{*}{4} 
& $23.68 \pm 1.38$ & $32.15 \pm 1.88$ & $29.38 \pm 1.69$ & $-9.70 \pm 24.34$ & $31.26 \pm 1.58$ & $33.22 \pm 1.65$ & $\mathbf{33.54 \boldsymbol{\pm} 1.73}$\\
& & & $0.66 \pm 0.03$ & $0.90 \pm 0.03$ & $0.86 \pm 0.02$ & $0.21 \pm 0.24$ & $0.89 \pm 0.02$ & $\mathbf{0.91 \boldsymbol{\pm} 0.02}$ & $\mathbf{0.91 \boldsymbol{\pm} 0.02}$\\
& & & $10.83 \pm 1.75$ & $1.44 \pm 0.58$ & $2.50 \pm 0.63$ & $126.49 \pm 296.11$ & $1.75 \pm 0.40$ & $1.11 \pm 0.33$ & $\mathbf{1.03 \boldsymbol{\pm} 0.34}$\\
\midrule
\multirow{3}{*}{R}
& \multirow{3}{*}{11} & \multirow{3}{*}{-} 
& $25.96 \pm 1.00$ & $33.42 \pm 1.16$ & $16.33 \pm 1.60$ & $-7.69 \pm 30.15$ & $33.25 \pm 1.13$ & $34.14 \pm 1.02$ & $\mathbf{34.30 \boldsymbol{\pm} 1.22}$\\
& & & $0.74 \pm 0.02$ & $\mathbf{0.91 \boldsymbol{\pm} 0.02}$ & $0.43 \pm 0.04$ & $0.28 \pm 0.30$ & $\mathbf{0.91 \boldsymbol{\pm} 0.02}$ & $\mathbf{0.91 \boldsymbol{\pm} 0.02}$ & $\mathbf{0.91 \boldsymbol{\pm} 0.02}$\\
& & & $6.32 \pm 0.82$ & $1.03 \pm 0.21$ & $24.34 \pm 5.45$ & $252.43 \pm 1194.97 $ & $1.06 \pm 0.21$ & $0.87 \pm 0.17$ & $\mathbf{0.84 \boldsymbol{\pm} 0.20}$\\
\midrule
\multirow{3}{*}{G}
& \multirow{3}{*}{8} & \multirow{3}{*}{-} 
& $31.38 \pm 0.86$ & $34.19 \pm 0.91$ & $22.67 \pm 1.84$ & $24.37 \pm 3.01$ & $33.84 \pm 0.81$ & $34.98 \pm 0.92$ & $\mathbf{35.04 \boldsymbol{\pm} 0.98}$\\
& & & $0.85 \pm 0.03$ & $\mathbf{0.92 \boldsymbol{\pm} 0.02}$ & $0.61 \pm 0.05$ & $0.75 \pm 0.05$ & $0.91 \pm 0.02$ & $\mathbf{0.92 \boldsymbol{\pm} 0.02}$ & $\mathbf{0.92 \boldsymbol{\pm} 0.02}$\\
& & & $1.71 \pm 0.33$ & $0.86 \pm 0.17$ & $9.43 \pm 2.35$ & $18.59 \pm 18.28$ & $0.94 \pm 0.18$ & $0.72 \pm 0.15$ & $\mathbf{0.71 \boldsymbol{\pm} 0.15}$\\
\midrule
\multicolumn{3}{c}{\multirow{2}{*}{\footnotesize \makecell{Number of\\Parameters}}} 
& \multirow{2}{*}{-} & \multirow{2}{*}{-} & \multirow{2}{*}{$3\times 10^{7}$} & \multirow{2}{*}{$447490$} & \multirow{2}{*}{$1558$} & \multirow{2}{*}{$1760$} & \multirow{2}{*}{$7348$} \\
& & & & & & & & & \\
\bottomrule
\end{tabular*}
\footnotetext{\textsuperscript{a} Rotated phase-encoding direction.}
\footnotetext{T: k-space Trajectory, C: Cartesian, R: Radial, G: 2D Gaussian, A: Acceleration, ACL: Auto Calibration Lines}
\label{tab:pi_braint2}
\end{table}

\begin{figure*}
	\centering
	\includegraphics{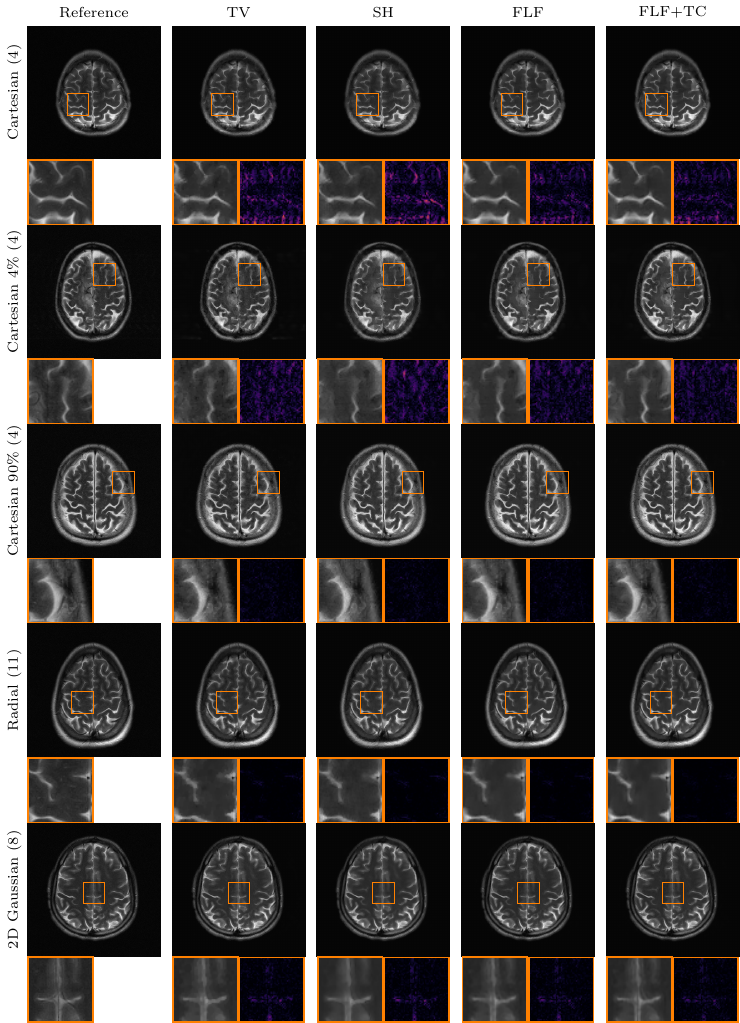}
	\caption{Qualitative \gls{pi} comparison on the brain T2 dataset against \gls{tv}.
	First row: 4-fold Cartesian k-space trajectory with $8\%$ available \glspl*{acl} and an acceleration of 4. Second row: Same as row one with $4\%$ available \glspl*{acl}. Third row: Same as row one with rotated phase-encoding direction. Fourth row: 2D Gaussian k-space trajectory with an acceleration of 8. Fifth row: Radial k-space trajectory with
   	an acceleration of 11. The inset shows a zoom on the image with corresponding absolute error. Our methods: Shearlet (SH), fully-learned filters (FLF) and fully-learned filters and time conditioning (FLF+TC).}
	\label{fig:qualitative_brain_T2}
\end{figure*}

The results for T2-weighted brain images show a similar trend, with our top-performing model surpassing \gls{tv} by 1 dB on average.
Interestingly, the shearlet model drops below \gls{tv} in terms of \gls{psnr} for this contrast, which we attribute to a higher noise level in the data. 
Both models with fully-learned convolutional filters remain robust across all k-space trajectories, where the learned time conditioning provides an additional 0.2 dB improvement on average.
The corresponding qualitative reconstructions are shown in~\Cref{fig:qualitative_brain_T2}. Here we highlight the Gaussian k-space trajectory in the last row, where we clearly observe typical staircasing artifacts in the \gls{tv} reconstruction. Furthermore, neither \gls{tv} nor the shearlet model recover the vertical edge separating the two brain hemispheres shown in the inset of the last row. By comparison, the models with fully-learned filters (and time conditioning) preserve this structural detail with higher clarity.

\subsubsection{Uncertainty Quantification}
\label{ssec:uncertainty_quantification}
In this section, we qualitatively evaluate the uncertainty quantification of our framework. The proposed method (\Cref{alg:joint_recon}) implements a stochastic map \(
\mathbf{y} \mapsto \tilde p_{\mathbf{X}, \mathbf{\Sigma} \mid \mathbf{Y}=\mathbf{y}}\), that transforms undersampled measurements into a conditional distribution of 
reconstructions and coil sensitivities. From samples of this distribution, we compute pixel-wise variance maps to represent local uncertainty in the reconstructed images.
We want to highlight, that due to algorithmic heuristics and the intractability of the true likelihood, \(\tilde p\) does not represent the true posterior distributions \(p_{\mathbf{X}, \mathbf{\Sigma} \mid \mathbf{Y}=\mathbf{y}}\), and is an approximation. Nonetheless, these variance maps remain a useful tool for analyzing the spatial distribution of uncertainty across different image regions.
\begin{figure*}
	\includegraphics{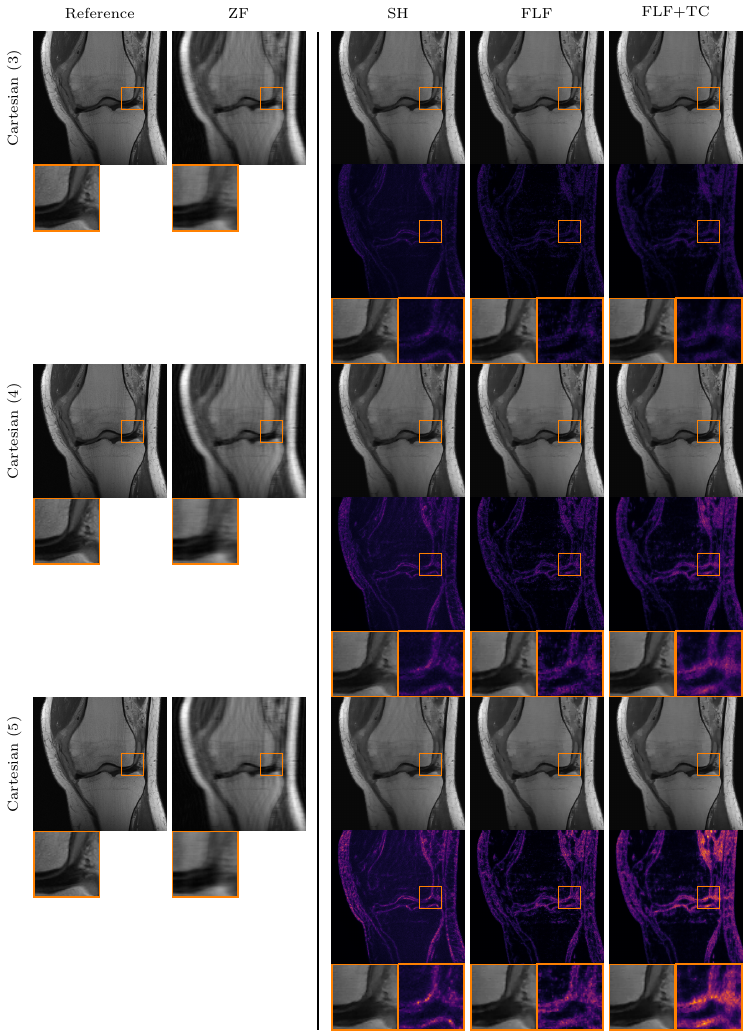}
    \caption{Qualitative comparison of pixel-wise variance maps for the shearlet model, model with fully-learned filters and the model with fully-learned filters and time conditioning for a Cartesian k-space trajectory with 8\% available \glspl{acl} and increasing acceleration factors between three and five. The left column shows the \gls{rss} reference, and the zero-filled reconstruction. In the right columns, the rows alternate between the Monte Carlo estimates of the approximate posterior expectation reconstructions for the different methods and their corresponding pixel-wise variance maps. Our methods: Shearlet (SH), fully-learned filters (FLF) and fully-learned filters and time conditioning (FLF+TC).}
	\label{fig:uncertainty_cartesian}
\end{figure*}
We demonstrate this in the prototypical setting of a Cartesian k-space trajectory with 8 \% available \glspl{acl} and increasing acceleration factors between three and five in~\Cref{fig:uncertainty_cartesian}. Specifically, we compute pixel-wise variance maps from 100 reconstructions for the shearlet model and the two models with fully-learned convolutional filters. As expected, uncertainty increases alongside the acceleration factor due to the reduction in available data. This is particularly evident in regions with fine structural details, such as blood vessels in fat tissue, which become increasingly difficult to reconstruct. Furthermore, the shearlet model exhibits slightly higher variance in homogeneous regions, such as the femur and tibia, relative to the other two~\gls{pogmdm} parameterizations. Finally, the model with learned time conditioning yields the highest pixel-wise variance across all acceleration factors. This qualitative observation is consistent with the quantitative results in \Cref{tab:pi_corpd}, where the model with learned time conditioning shows slightly larger standard deviations across all comparison metrics.

\begin{figure*}[!ht]
    \centering
	\includegraphics{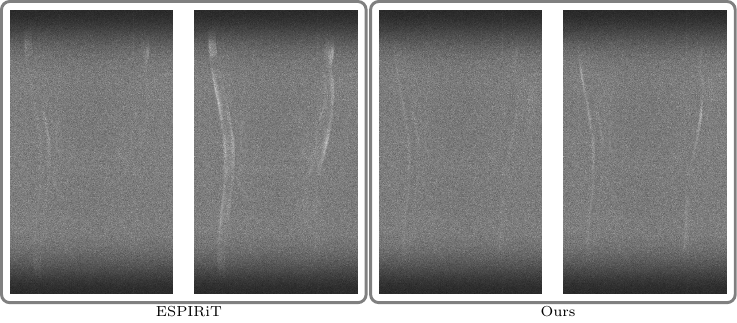}
    \caption{%
		\gls{rss} nullspace residuals for Cartesian k-space trajectories.
		In each block, the measured data has 8\% available \glspl*{acl} (left) and 4\% available \glspl*{acl} (right).%
	}%
    \label{fig:rss_nullspace}
\end{figure*}

\subsubsection{Coil Sensitivities}
\label{ssec:coil_sensitivities}
In this section we shift the focus to the estimated coil sensitivities of the joint reconstruction. Here, we compare the estimated coil sensitivities qualitatively to ESPIRiT~\cite{uecker2014espirit} by investigating their \gls{rss} null-space residuals. The null-space residual of the \(i\)-th coil is given by
\begin{equation}
\label{eq:null_space_residuals}
\frac{\mathbf{s}_i}{|\mathbf{s}|^2_{\text{RSS}}} \left ( \sum_{j=1}^c \bar{\mathbf{s}}_j\mathbf{x}_j \right ) - \mathbf{x}_i,
\end{equation}
where \(|\cdot|_{\text{RSS}} \) implements the map \( \mathbf{s} \mapsto \sqrt{\sum_{i=1}^c |\mathbf{s}_i|^2} \). Ideally, the residuals should only contain noise when the estimated coil sensitivities are exact.
Similar to~\cite{zach2023stabledeepmri,erlacher2023jointnonlinearmriinversion}, we compare the \gls{rss} reduction of the individual coil residuals, as shown in~\Cref{fig:rss_nullspace}. While~\cite{uecker2014espirit} provides slightly better coil sensitivity estimates when a larger number of \glspl{acl} is available, its performance degrades as the calibration region decreases. By contrast, our method delivers robust coil sensitivity estimates with less available data in the k-space center.

\subsection{Ablation Study}
\label{ssec:ablation}
We now turn our focus on the influence of different parameter choices of the shearlet model and show numerical results for more efficient parameterization. Due to computational constraints, quantitative results in this section are generated by approximating the posterior expectation with a single reconstruction. \Cref{tab:shear_levels} presents results for different number of scalings \( j \) of the shearlet system. We observe a substantial improvement when increasing the scaling from \( j = 1 \) to \( j = 2 \). This is expected, as it allows the model to capture a broader range of directional components, thus improving its ability to represent fine structural details such as edges. Increasing to three scales yields minor improvements, while four scales slightly degrade reconstruction quality. We note that further hyperparameter tuning at \( j= 4 \) might mitigate this. The results support the choice of a scaling of two, as it balances downstream performance and inference speed which grows with higher scaling.

\begin{table}[t]
\centering
\caption{Quantitative \gls{pi} reconstruction results for different shearlet scalings \( j \) on the \gls{corpd} dataset. 
The rows alternate between \gls{psnr}, \gls{ssim} and \gls{nmse}. The \gls{nmse} is scaled by $10^2$. All metrics are shown as mean $\pm$ unit standard deviation. Bold typeface indicates the best method. Comparison method: Zero filled (ZF).}
\begin{tabular*}{\textwidth}{@{\extracolsep\fill}cccccccc} 
\toprule
\multirow{2}{*}{T} & \multirow{2}{*}{A} & \multirow{2}{*}{ACL} & \multirow{2}{*}{ZF} & \multicolumn{4}{c}{Shearlet}\\
\cmidrule(lr){5-8}
& & & & $j=1$ & $j=2$ & $j=3$ & $j=4$ \\
\midrule
\multirow{9}{*}{C}
& \multirow{9}{*}{4} & \multirow{3}{*}{8} 
& $27.19 \pm 1.85$ & $33.65 \pm 1.74$ & $\mathbf{34.06 \boldsymbol{\pm} 1.81} $ & $34.04 \pm 1.82$ & $33.95 \pm 1.90$ \\
& & & $0.74 \pm 0.04$ & $0.86 \pm 0.03$ & $\mathbf{0.87 \boldsymbol{\pm} 0.03}$ & $\mathbf{0.87 \boldsymbol{\pm} 0.03}$ & $0.86 \pm 0.03$\\
& & & $2.37 \pm 0.72$ & $0.51 \pm 0.14$ & $\mathbf{0.46 \boldsymbol{\pm} 0.14}$ & $\mathbf{0.46 \boldsymbol{\pm} 0.13}$ & $0.47 \pm 0.14$ \\
\cmidrule(l){3-8}
& & \multirow{3}{*}{8\footnotemark[1]} 
& $31.13 \pm 2.36$ & $34.82 \pm 1.76$ & $35.27 \pm 1.85$ & $\mathbf{35.30 \boldsymbol{\pm} 1.71}$ & $35.15 \pm 1.73$ \\
& & & $0.81 \pm 0.05$ & $0.88 \pm 0.03$ & $\mathbf{0.89 \boldsymbol{\pm} 0.03}$ & $\mathbf{0.89 \boldsymbol{\pm} 0.03} $ & $\mathbf{0.89 \boldsymbol{\pm} 0.03}$ \\
& & & $0.97 \pm 0.40$ & $0.39 \pm 0.11$ & $\mathbf{0.35 \boldsymbol{\pm} 0.11}$ & $\mathbf{0.35 \boldsymbol{\pm} 0.09}$ & $0.36 \pm 0.09$ \\
\cmidrule(l){3-8}
& & \multirow{3}{*}{4} 
& $24.14 \pm 1.66$ & $33.52 \pm 1.80$ & $33.92 \pm 1.73$ & $33.92 \pm 1.75$ & $\mathbf{33.97 \boldsymbol{\pm} 1.85}$ \\
& & & $0.69 \pm 0.04$ & $\mathbf{0.87 \boldsymbol{\pm} 0.03}$ & $\mathbf{0.87 \boldsymbol{\pm} 0.03}$ & $\mathbf{0.87 \boldsymbol{\pm} 0.03}$ & $\mathbf{0.87 \boldsymbol{\pm} 0.03}$ \\
& & & $4.96 \pm 1.52$ & $0.53 \pm 0.15$ & $0.48 \pm 0.13$ & $0.48 \pm 0.12$ & $\mathbf{0.47 \boldsymbol{\pm} 0.13}$\\
\midrule
\multirow{3}{*}{R} 
& \multirow{3}{*}{11} & \multirow{3}{*}{-}
& $28.76 \pm 2.15$ & $32.79 \pm 1.86$ & $33.68 \pm 1.80$ & $\mathbf{33.74 \boldsymbol{\pm} 1.77}$ & $33.55 \pm 1.81$ \\
& & & $0.75 \pm 0.06$ & $0.83 \pm 0.04$ & $\mathbf{0.85 \boldsymbol{\pm} 0.03}$ & $\mathbf{0.85 \boldsymbol{\pm} 0.03}$ & $0.84 \pm 0.03$ \\
& & & $1.67 \pm 0.58$ & $0.62 \pm 0.17$ & $0.50 \pm 0.14$ & $\mathbf{0.49 \boldsymbol{\pm} 0.13}$ & $0.52 \pm 0.14$ \\
\midrule
\multirow{3}{*}{G} 
& \multirow{3}{*}{8} & \multirow{3}{*}{-} 
& $ 32.15 \pm 2.32$ & $33.44 \pm 1.91$ & $34.47 \pm 1.91$ & $\mathbf{34.53 \boldsymbol{\pm} 1.89}$ & $34.33 \pm 1.94$ \\
& & & $0.84 \pm 0.05$ & $0.85 \pm 0.04$ & $\mathbf{0.87 \boldsymbol{\pm} 0.04}$ & $\mathbf{0.87 \boldsymbol{\pm} 0.03}$ & $0.86 \pm 0.04$ \\
& & & $0.78 \pm 0.35$ & $0.53 \pm 0.15$ & $\mathbf{0.42 \boldsymbol{\pm} 0.12}$ & $\mathbf{0.42 \boldsymbol{\pm} 0.12}$ & $0.44 \pm 0.13$ \\
\midrule
\multicolumn{3}{c}{\multirow{2}{*}{\footnotesize \makecell{Number of\\Parameters}}} & \multirow{2}{*}{-} & \multirow{2}{*}{$928$} & \multirow{2}{*}{$1558$} & \multirow{2}{*}{$2962$} & \multirow{2}{*}{$3826$} \\
& & & & & & & \\
\bottomrule
\end{tabular*}
\footnotetext{\textsuperscript{a} Rotated phase-encoding direction.}
\footnotetext{T: k-space Trajectory, C: Cartesian, R: Radial, G: 2D Gaussian, A: Acceleration, ACL: Auto Calibration Lines}
\label{tab:shear_levels}
\end{table}

In~\Cref{ssec:learned_models}, we observed similarities in the learned potentials of the shearlet model (see the potentials in~\Cref{fig:learned_model}). To exploit these, we consider three sharing strategies: (i) sharing potentials within the vertical and horizontal cone at a given shearing scale \( j \) (ii) sharing potentials across the vertical and horizontal cone at the same scale \(j\) and (iii) combining both (i) and (ii). For instance, in strategy (i) we share the first and fifth potential in the first row in~\Cref{fig:learned_model}, in (ii) we share the first and second row and in (iii) we share both within and across the row reducing the number of different potentials per scale to three. This reduces the total number of parameters from \( 1558 \) to \( 676 \). The quantitative results in~\Cref{tab:shearlet_shared_variations} reveal that for all sharing strategies, the performance in terms of \gls{psnr} stays within \( \pm 0.12\) dB of the shearlet baseline without sharing. Interestingly, the configuration with the lowest parameter count slightly outperforms the baseline on the Gaussian k-space trajectory, which may suggest that the grid search on the limited validation set did not identify optimal hyperparameters here. However, we emphasize that the results in~\Cref{tab:shearlet_shared_variations} are based on a single reconstruction.

\begin{table}[t]
\centering
\caption{Influence of different sharing strategies of the Gaussian mixture factors on the \gls{corpd} dataset.  
The rows alternate between \gls{psnr}, \gls{ssim} and \gls{nmse}. The \gls{nmse} is scaled by $10^2$. All metrics are shown as mean $\pm$ unit standard deviation. Bold typeface indicates the best method.}
\begin{tabular*}{\textwidth}{@{\extracolsep\fill}cccccccc} 
\toprule
\multirow{2}{*}{T} & \multirow{2}{*}{A} & \multirow{2}{*}{\gls{acl}} & \multirow{2}{*}{ZF} & \multirow{2}{*}{Shearlet} & \multicolumn{3}{c}{Shared} \\
\cmidrule(lr){6-8}
& & & &  & Within & Across & Both \\
\midrule
\multirow{9}{*}{C}
& \multirow{9}{*}{4} & \multirow{3}{*}{8} 
& $27.19 \pm 1.85$ & $\mathbf{34.06 \boldsymbol{\pm} 1.81}$  & $34.05 \pm 1.81$ & $34.04 \pm 1.76$ & $34.05 \pm 1.77$ \\
& & & $0.74 \pm 0.04$ & $\mathbf{0.87 \boldsymbol{\pm} 0.03}$ & $\mathbf{0.87 \boldsymbol{\pm} 0.03}$ & $\mathbf{0.87 \boldsymbol{\pm} 0.03}$ & $\mathbf{0.87 \boldsymbol{\pm} 0.03}$ \\
& & & $2.37 \pm 0.72$ & $\mathbf{0.46 \boldsymbol{\pm} 0.14}$ & $\mathbf{0.46 \boldsymbol{\pm} 0.14}$ & $\mathbf{0.46 \boldsymbol{\pm} 0.13}$ & $\mathbf{0.46 \boldsymbol{\pm} 0.13}$ \\
\cmidrule(l){3-8}
& & \multirow{3}{*}{8\footnotemark[1]} 
& $31.13 \pm 2.36$ & $35.27 \pm 1.85$ & $35.25 \pm 1.85 $ & $\mathbf{35.32 \boldsymbol{\pm} 1.85}$ & $\mathbf{35.32 \boldsymbol{\pm} 1.85}$ \\
& & & $0.81 \pm 0.05$ & $\mathbf{0.89 \boldsymbol{\pm} 0.03}$ & $\mathbf{0.89 \boldsymbol{\pm} 0.03}$ & $\mathbf{0.89 \boldsymbol{\pm} 0.03}$ & $\mathbf{0.89 \boldsymbol{\pm} 0.03}$ \\
& & & $0.97 \pm 0.40$ & $\mathbf{0.35 \boldsymbol{\pm} 0.11}$ & $\mathbf{0.35 \boldsymbol{\pm} 0.11}$ & $\mathbf{0.35 \boldsymbol{\pm} 0.11}$ & $\mathbf{0.35 \boldsymbol{\pm} 0.11}$ \\
\cmidrule(l){3-8}
& & \multirow{3}{*}{4} 
& $24.14 \pm 1.66$ & $33.92 \pm 1.73$ & $\mathbf{33.93 \boldsymbol{\pm} 1.72}$  & $33.84 \pm 1.66$ & $33.83 \pm 1.67$ \\
& & & $0.69 \pm 0.04$ & $\mathbf{0.87 \boldsymbol{\pm} 0.03}$ & $\mathbf{0.87 \boldsymbol{\pm} 0.03}$ & $\mathbf{0.87 \boldsymbol{\pm} 0.03}$ & $\mathbf{0.87 \boldsymbol{\pm} 0.03}$\\
& & & $4.96 \pm 1.52$ & $\mathbf{0.48 \boldsymbol{\pm} 0.13}$ & $\mathbf{0.48 \boldsymbol{\pm} 0.13}$ & $0.49 \pm 0.12$ & $0.49 \pm 0.12$ \\
\midrule
\multirow{3}{*}{R} 
& \multirow{3}{*}{11} & \multirow{3}{*}{-} 
& $28.76 \pm 2.15$ & $33.68 \pm 1.80$ & $33.67 \pm 1.79$ &  $33.73 \pm 1.79$ & $\mathbf{33.74 \boldsymbol{\pm} 1.80}$ \\
& & & $0.75 \pm 0.06$ & $\mathbf{0.89 \boldsymbol{\pm} 0.03}$ & $\mathbf{0.89 \boldsymbol{\pm} 0.03}$ & $\mathbf{0.89 \boldsymbol{\pm} 0.03}$ & $\mathbf{0.89 \boldsymbol{\pm} 0.03}$ \\
& & & $1.67 \pm 0.58$ & $\mathbf{0.50 \boldsymbol{\pm} 0.14}$ & $\mathbf{0.50 \boldsymbol{\pm} 0.14}$ &  $\mathbf{0.50 \boldsymbol{\pm} 0.13}$ & $\mathbf{0.50 \boldsymbol{\pm} 0.13}$ \\
\midrule
\multirow{3}{*}{G} 
& \multirow{3}{*}{8} & \multirow{3}{*}{-} 
& $ 32.15 \pm 2.32$ & $34.47 \pm 1.91$ & $34.46 \pm 1.90$ &  $\mathbf{35.59 \boldsymbol{\pm} 1.92}$ & $\mathbf{34.59 \boldsymbol{\pm} 1.92}$ \\
& & & $0.84 \pm 0.05$ & $\mathbf{0.87 \boldsymbol{\pm} 0.04}$ & $\mathbf{0.87 \boldsymbol{\pm} 0.04}$ &  $\mathbf{0.87 \boldsymbol{\pm} 0.03}$ & $\mathbf{0.87 \boldsymbol{\pm} 0.03}$ \\
& & & $0.78 \pm 0.35$ & $0.42 \pm 0.12$ & $0.42 \pm 0.12$ &  $\mathbf{0.41 \boldsymbol{\pm} 0.12}$ & $\mathbf{0.41 \boldsymbol{\pm} 0.12}$ \\
\midrule
\multicolumn{3}{c}{\multirow{2}{*}{\footnotesize \makecell{Number of\\Parameters}}} & \multirow{2}{*}{-} & \multirow{2}{*}{$1558$} & \multirow{2}{*}{$928$} & \multirow{2}{*}{$928$} & \multirow{2}{*}{$676$} \\
& & & & & & & \\
\bottomrule
\end{tabular*}
\footnotetext{\textsuperscript{a} Rotated phase-encoding direction.}
\footnotetext{T: k-space Trajectory, C: Cartesian, R: Radial, G: 2D Gaussian, A: Acceleration, ACL: Auto Calibration Lines}
\label{tab:shearlet_shared_variations}
\end{table}
In~\Cref{ssec:alternative_pogmdm_parametrizations} we introduce the parameterization of the time-embedding network \( \tau_\theta \) as a sequence of linear layers and \gls{elu} activations followed by a final softplus activation at the output. Interestingly, most of the learned time conditionings in~\Cref{fig:output_time_embedding} follow simple functional forms. Therefore, we propose to adapt the variances of the \(k\)-th one-dimensional Gaussian mixture factor as
\begin{equation}
	\label{eq:softplus_parametrization}
	\sigma_{k}^2(t) = \sigma_0^2 + (\hat \tau_\theta(t))_k
\end{equation}
where 
\begin{equation}
	\label{eq:softplus_parametrization_network}
	(\hat{\tau}_{\theta}(t))_k = \theta_{1,k} \text{softplus}(\theta_{2,k} \sqrt{2t} + \theta_{3,k}), \quad k = 1, \dots o,
\end{equation}
and \( \theta = (\theta_{1,k}, \theta_{2,k}, \theta_{3,k})\) for \( k = 1,\dots,o\) denote learnable parameters. We ensure strictly positive outputs by constraining \( \theta_{1,1},\dots,\theta_{1,o}  > 0\). This reduces the number of learnable parameters from \( 7348 \) to \( 1820 \).
\begin{table}[!h]
	\centering
	\caption{Quantitative results on \gls{corpd} for different parameterizations of the learned time conditioning. The rows alternate between \gls{psnr}, \gls{ssim} and \gls{nmse}. The \gls{nmse} is scaled by $10^2$. All metrics are shown as mean $\pm$ unit standard deviation. Bold typeface indicates the best method. We denote the k-space trajectories as their abbreviation followed by the acceleration and the number of \glspl{acl} for Cartesian k-space trajectories.}
	\begin{tabular*}{\textwidth}{@{\extracolsep\fill}ccccccc}
	\toprule
	\multirow{2}{*}{\footnotesize \makecell{Time\\Conditioning}} & \multicolumn{5}{c}{T} & \multirow{2}{*}{\footnotesize \makecell{Number of\\Parameters}} \\
	\cmidrule(lr){2-6}
	& C(4/8\% \gls{acl}) & C\footnotemark[1](4/8\% \gls{acl})  & C(4/4\% \gls{acl}) & R(11) & G(8) \\
	\midrule
	\multirow{3}{*}{\eqref{eq:learned_embedding}}
	& \(\mathbf{34.69 \boldsymbol{\pm} 2.61}\) & \(\mathbf{36.11 \boldsymbol{\pm} 2.21}\) & \(\mathbf{34.51 \boldsymbol{\pm} 2.29}\) & \(\mathbf{34.24 \boldsymbol{\pm} 2.27}\) & \(\mathbf{35.08 \boldsymbol{\pm} 2.09}\) & \multirow{3}{*}{7348} \\
	& \(\mathbf{0.88 \boldsymbol{\pm} 0.04}\)   & \(\mathbf{0.90 \boldsymbol{\pm} 0.03}\)   & \(\mathbf{0.88 \boldsymbol{\pm} 0.04}\)   & \( \mathbf{0.86 \boldsymbol{\pm} 0.04} \) & \( \mathbf{0.88 \boldsymbol{\pm} 0.04} \)  \\
	& \(\mathbf{0.44 \boldsymbol{\pm} 0.39}\)   & \(\mathbf{0.30 \boldsymbol{\pm} 0.12}\)   & \(\mathbf{0.45 \boldsymbol{\pm} 0.29}\)   & \(\mathbf{0.46 \boldsymbol{\pm} 0.19}\) & \(\mathbf{0.37 \boldsymbol{\pm} 0.13}\)  \\
	\midrule
	\multirow{3}{*}{\eqref{eq:softplus_parametrization}}
	& \(34.35 \pm 2.13\) & \(35.41 \pm 1.58\) & \(34.42 \pm 2.03\) & \(34.02 \pm 1.75\)& \(34.97 \pm 1.83\) & \multirow{3}{*}{1820}\\
	& \(0.87 \pm 0.03\)   & \(0.89 \pm 0.02\)   & \(0.87 \pm 0.03\)   & \(0.85 \pm 0.03\)  & \(\mathbf{0.88 \boldsymbol{\pm} 0.03}\) \\
	& \(0.46 \pm 0.37\)   & \(0.34 \pm 0.09\)   & \(0.44 \pm 0.24\)   & \(0.47 \pm 0.13\)  & \(0.38 \pm 0.11\) \\
	\bottomrule
	\end{tabular*}
	\footnotetext{\textsuperscript{a} Rotated phase-encoding direction.}
	\footnotetext{T: k-space Trajectory, C: Cartesian, R: Radial, G: 2D Gaussian, A: Acceleration, ACL: Auto Calibration Lines}
	\label{tab:tau_theta_simple}
\end{table}
 We compare the model with fully-learned filters and time conditioning according to \eqref{eq:learned_embedding} with the model with fully-learned filters and the time conditioning as in \eqref{eq:softplus_parametrization} in \Cref{tab:tau_theta_simple}.
Here we observe that the model with the time conditioning according to~\eqref{eq:softplus_parametrization} is around \( 0.2 \) dB worse for all k-space trajectories except the Cartesian one with rotated phase-encoding direction, where the drop is \( 0.6 \) dB. We did not adapt any hyperparameters between the two cases which may explain the drop in performance here. Furthermore, we again emphasize that the presented metrics are based on a single reconstruction. 
\section{Conclusion}\label{sec:conclusion}
In this work, we combined a product-of-Gaussian-mixture diffusion model with a classical smoothness penalty to tackle the nonlinear magnetic resonance imaging reconstruction problem within a diffusion-based framework. The shearlet model achieves good performance in both single- and multi-coil settings while maintaining fast inference. We improve upon the shearlet baseline by introducing additional degrees of freedom through fully-learned filters and time conditioning in the model. The proposed approach demonstrates robustness to shifts in the underlying anatomy, the contrast mechanism, and can be made more parameter-efficient by sharing potentials or learning simpler functional forms for the time conditioning. Limitations of our approach are the need to tune multiple hyperparameters, deviations from the conditional reverse diffusion, suboptimal training and limited reconstruction quality. Interesting future directions include extending the diffusion prior to the joint time-dependent density \( p_{\mathbf{X}_t,\mathbf{\Sigma}_t} \) to align more closely with recent posterior-sampling algorithms. Additionally, the diffusion prior can be made more expressive by learning the time conditioning not only per-factor but per-component in the one-dimensional Gaussian mixture factors, or by designing a multi-layer product-of-Gaussian-mixture diffusion model. 

\section*{Declarations}
\textbf{Author contribution} L.N., M.Z., and T.P. contributed to the conceptualization. L.N. conducted the experiments and prepared the first draft of the manuscript. M.Z. and T.P. reviewed the draft. The final manuscript was written by L.N. and M.Z.\\

\noindent
\textbf{Funding} This research was funded in whole or in part by the Austrian Science
Fund (FWF) \href{https://dx.doi.org/10.55776/F100800}{10.55776/F100800}.

\backmatter

\begin{appendices}
\section{Noise Schedule}\label{sec:appendix_A}%

Exponential noise schedules are commonly used as discretization points of the reverse-time \gls{sde} in score-based diffusion models~\cite{erlacher2023jointnonlinearmriinversion,song2021scorebased,chung2022scoreMRI,luo2023bayesianmri}. A general formulation is given by
\begin{equation}
\label{eq:noise_schedule}
	\zeta(t) = \zeta_{\max}\left(\frac{\zeta_{\min}}{\zeta_{\max}}\right)^{(1 - t/T)^p},
\end{equation}
with $p \in \mathbb{R}_+$.

\begin{figure}
	\centering
	\includegraphics{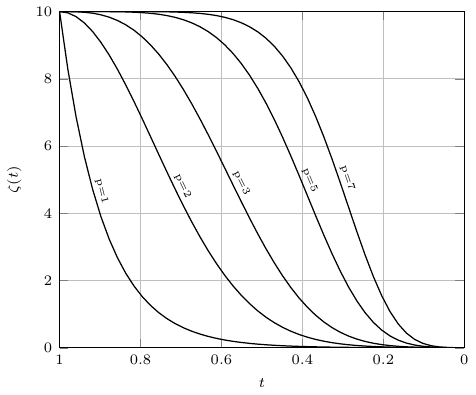}
	\caption{Noise schedules defined in~\eqref{eq:noise_schedule} for different parameters $p$. Larger values of $p$ keep the schedule at higher noise levels for a longer duration before decaying more rapidly. For $p=1$, the schedule remains close to zero as $t \to 0$ for a significant part of the schedule.}
\label{fig:noise_schedule}
\end{figure}

\Cref{fig:noise_schedule} illustrates schedules for different values of $p$, using $\zeta_{\min}=0.001$, $\zeta_{\max}=10$ and $T = 1$. We observe that the schedule for $p=1$ decays fast and then stays close to zero for a big part of the reverse diffusion. In contrast, for higher $p$, the schedule stays at a higher noise level for a longer period and then decays rapidly towards zero. Importantly, \eqref{eq:noise_schedule} corresponds to a geometric progression only when \( p=1 \) a case popularized in~\cite{song2020trainingsbms}. Since we employ the acceleration method proposed in~\cite{chung2022ccdf}, we found it advantageous to choose $p=5$ as we start from $t' = 0.2T$ in the reverse diffusion. This choice allows us to leverage the learned model at higher diffusion times while still benefiting from the accelerated reverse process.

\section{Derivation of Tweedie's Formula}\label{sec:appendix_tweedie}
In the following we derive Tweedie's formula. Our starting point is the expression in~\eqref{eq:solution_heat_equation} which states
\begin{align*}
	p_{\mathbf{X}_t}(\mathbf{x}) = N_{0, 2tI}(\mathbf{x}) * p_{\mathbf{X}_0}(\mathbf{x}) = \int N_{0, 2t\mathbf{I}}(\mathbf{x} - \hat{\mathbf{x}})p_{\mathbf{X}_0}(\hat{\mathbf{x}})d\hat{\mathbf{x}} 
\end{align*}
Taking the gradient w.r.t. \( \mathbf{x} \) then leads to
\begin{align*}
	\nabla p_{\mathbf{X}_t}(\mathbf{x}) &= \int -\frac{(\mathbf{x} - \hat{\mathbf{x}})}{2t}N_{0, 2t\mathbf{I}}(\mathbf{x} - \hat{\mathbf{x}})p_{\mathbf{X}_0}(\hat{\mathbf{x}})d\hat{\mathbf{x}} \\
	&= -\frac{1}{2t} \left ( \mathbf{x} \int N_{0, 2t\mathbf{I}}(\mathbf{x} - \hat{\mathbf{x}})p_{\mathbf{X}_0}(\hat{\mathbf{x}})d\hat{\mathbf{x}} - \int \hat{\mathbf{x}}N_{0, 2t\mathbf{I}}(\mathbf{x} - \hat{\mathbf{x}})p_{\mathbf{X}_0}(\hat{\mathbf{x}})d\hat{\mathbf{x}} \right ) \\
	&= -\frac{1}{2t} \big( \mathbf{x}p_{\mathbf{X}_t}(\mathbf{x}) - p_{\mathbf{X}_t}(\mathbf{x})\mathbb{E}[\mathbf{X}_0 \mid \mathbf{X}_t = \mathbf{x}] \big).
\end{align*}
Where we use the definition of the conditional expectation 
\begin{equation*}
	\mathbb{E}[\mathbf{X}_0 \mid \mathbf{X}_t = \mathbf{x}] = \int \hat{\mathbf{x}} p_{\mathbf{X}_0\mid \mathbf{X}_t}(\hat{\mathbf{x}}\mid \mathbf{x}) \mathrm{d}\hat{\mathbf{x}} = \frac{1}{p_{\mathbf{X}_t}(\mathbf{x})} \int \hat{\mathbf{x}} p_{\mathbf{X}_t\mid \mathbf{X}_0}(\mathbf{x}\mid \hat{\mathbf{x}})p_{\mathbf{X}_0}(\hat{\mathbf{x}}) \mathrm{d}\hat{\mathbf{x}}.
\end{equation*}
Finally, dividing by \( p_{\mathbf{X}_t}(\mathbf{x}) \) yields
\begin{equation*}
	\frac{\nabla p_{\mathbf{X}_t}(\mathbf{x})}{p_{\mathbf{X}_t}(\mathbf{x})} = \nabla \log p_{\mathbf{X}_t}(\mathbf{x}) = -\frac{1}{2t} \big( \mathbf{x} - \mathbb{E}[\mathbf{X}_0 \mid \mathbf{X}_t = \mathbf{x}] \big),
\end{equation*}
which is equal to~\eqref{eq:tweedies_formula} when rearranging the terms accordingly.

\section{Denoising Experiment}\label{sec:appendix_B}
In this section, we present results for the denoising problem with increasing noise levels. The denoising problem with additive Gaussian noise can be formulated as
\begin{equation}
\label{eq:denoising_problem}
\mathbf{y} = \mathbf{x} + \sigma\mathbf{n},
\end{equation}
where \( \mathbf{n} \sim N_{0,\mathbf{I}} \) and \( \sigma \) denotes the noise level. The goal is to recover the clean signal $\mathbf{x}$ from the noisy measurement $\mathbf{y}$.
Specifically, we aim to estimate the posterior expected value \( \mathbf{y} \mapsto \mathbb{E}[\mathbf{X}|\mathbf{Y}=\mathbf{y}] = \int \mathbf{x}p_{\mathbf{X}|\mathbf{Y}}(\mathbf{x},\mathbf{y})d\mathbf{x} \). This map can be expressed using Tweedie's formula~\cite{robbins1956empiricalbayes,efron2011tweedie} which we used in~\eqref{eq:tweedies_formula}. 
We show quantitative denoising results on the \gls{corpd} dataset for shearlet models with scalings $j=1$ and $j=2$, as well as the model with fully-learned convolutional filters and time conditioning in \Cref{tab:denoising_results}.
\begin{table}[t]
\centering
\caption{Quantitative denoising results using one-step empirical Bayes denoising with different \gls{pogmdm} parameterizations. \gls{psnr} and \gls{ssim} are presented as mean $\pm$ unit standard deviation. Bold typeface indicates the best method. Our methods: Shearlet (SH), fully-learned filters (FLF) and fully-learned filters and time conditioning (FLF+TC). The number of shearlet scalings is denotes as \( j \).}
\begin{tabular*}{\textwidth}{@{\extracolsep\fill} c c c c c c c c c c}
\toprule
& & & $\sigma$ & $\mathbf{y}$ & SH ($j=1$) & SH ($j=2$) & FLF & FLF+TE & \\
\midrule
\multicolumn{2}{c}{\multirow{8}{*}{\rotatebox{90}{CORPD}}} &
\multicolumn{1}{c}{\multirow{4}{*}{\rotatebox{90}{PSNR}}} 
& $0.025$ & $32.04 \pm 0.02$ & $35.82 \pm 0.73$ & $36.06 \pm 0.79$ & $36.69 \pm 0.82$ & $\mathbf{36.97 \boldsymbol{\pm} 0.80}$ \\
& & & $0.050$ & $26.02 \pm 0.02$ & $32.62 \pm 1.00$ & $32.91 \pm 1.00$ & $33.83 \pm 1.00$ & $\mathbf{33.96 \boldsymbol{\pm} 1.02}$ \\
& & & $0.100$ & $20.00 \pm 0.02$ & $29.98 \pm 1.10$ & $30.24 \pm 1.08$ & $31.03 \pm 0.99$ & $\mathbf{31.24 \boldsymbol{\pm} 1.03}$ \\
& & & $0.200$ & $13.99 \pm 0.02$ & $27.31 \pm 0.85$ & $27.66 \pm 0.96$ & $27.99 \pm 0.93$ & $\mathbf{28.21 \boldsymbol{\pm} 0.91}$ \\
\cmidrule(l){3-10}
& & \multicolumn{1}{c}{\multirow{4}{*}{\rotatebox{90}{SSIM}}} 
& $0.025$ & $0.77 \pm 0.02$ & $0.90 \pm 0.00$ & $0.90 \pm 0.00$ & $0.91 \boldsymbol{\pm}  0.01$ & $\mathbf{0.92 \boldsymbol{\pm} 0.01}$ \\
& & & $0.050$ & $0.50 \pm 0.04$ & $0.82 \pm 0.01$ & $0.83 \pm 0.01$ & $\mathbf{0.85 \boldsymbol{\pm} 0.01}$ & $\mathbf{0.85 \boldsymbol{\pm} 0.01}$ \\
& & & $0.100$ & $0.25 \pm 0.04$ & $0.73 \pm 0.02$ & $0.74 \pm 0.02$ & $\mathbf{0.77 \boldsymbol{\pm} 0.02}$ & $\mathbf{0.77 \boldsymbol{\pm} 0.02}$ \\
& & & $0.200$ & $0.10 \pm 0.02$ & $0.59 \pm 0.02$ & $0.63 \pm 0.02$ & $0.64 \pm 0.02$ & $\mathbf{0.65 \boldsymbol{\pm} 0.02}$ \\
\midrule
\multicolumn{2}{c}{\multirow{8}{*}{\rotatebox{90}{CORPDFS}}} &
\multicolumn{1}{c}{\multirow{4}{*}{\rotatebox{90}{PSNR}}} 
& $0.025$ & $32.04 \pm 0.02$ & $35.21 \pm 0.97$ & $35.27 \pm 1.16$ & $35.70 \pm 1.16$ & $\mathbf{35.80 \boldsymbol{\pm} 1.13}$ \\
& & & $0.050$ & $26.02 \pm 0.02$ & $32.19 \pm 1.14$ & $32.36 \pm 1.23$ & $33.04 \pm 1.33$ & $\mathbf{33.16 \boldsymbol{\pm} 1.38}$ \\
& & & $0.100$ & $20.00 \pm 0.02$ & $29.79 \pm 1.07$ & $29.98 \pm 1.10$ & $30.57 \pm 1.20$ & $\mathbf{30.77 \boldsymbol{\pm} 1.26}$ \\
& & & $0.200$ & $13.98 \pm 0.02$ & $27.36 \pm 0.79$ & $27.72 \pm 0.87$ & $27.90 \pm 0.90$ & $\mathbf{28.03 \boldsymbol{\pm} 0.93}$ \\
\cmidrule(l){3-10}
& & \multicolumn{1}{c}{\multirow{4}{*}{\rotatebox{90}{SSIM}}} 
& $0.025$ & $0.78 \pm 0.04$ & $\mathbf{0.87 \boldsymbol{\pm} 0.01}$ & $\mathbf{0.87 \boldsymbol{\pm} 0.02}$ & $\mathbf{0.87 \boldsymbol{\pm} 0.02}$ & $\mathbf{0.87 \boldsymbol{\pm} 0.02}$ \\
& & & $0.050$ & $0.50 \pm 0.06$ & $0.78 \pm 0.04$ & $0.78 \pm 0.04$ & $\mathbf{0.79 \boldsymbol{\pm} 0.05}$ & $\mathbf{0.79 \boldsymbol{\pm} 0.05}$ \\
& & & $0.100$ & $0.24 \pm 0.04$ & $0.68 \pm 0.06$ & $0.69 \pm 0.06$ & $\mathbf{0.71 \boldsymbol{\pm} 0.07}$ & $\mathbf{0.71 \boldsymbol{\pm} 0.07}$ \\
& & & $0.200$ & $0.09 \pm 0.02$ & $0.56 \pm 0.05$ & $0.59 \pm 0.06$ & $0.60 \pm 0.06$ & $\mathbf{0.61 \boldsymbol{\pm} 0.07}$ \\
\midrule
\multicolumn{5}{c}{\multirow{2}{*}{\footnotesize \makecell{Number of\\Parameters}}} & \multirow{2}{*}{\footnotesize $928$} & \multirow{2}{*}{\footnotesize $1558$} & \multirow{2}{*}{\footnotesize $1760$} & \multirow{2}{*}{\footnotesize $7348$} & \\
& & & & & & & & & \\
\bottomrule
\end{tabular*}
\label{tab:denoising_results}
\end{table}

\begin{figure*}
	\centering
	\includegraphics{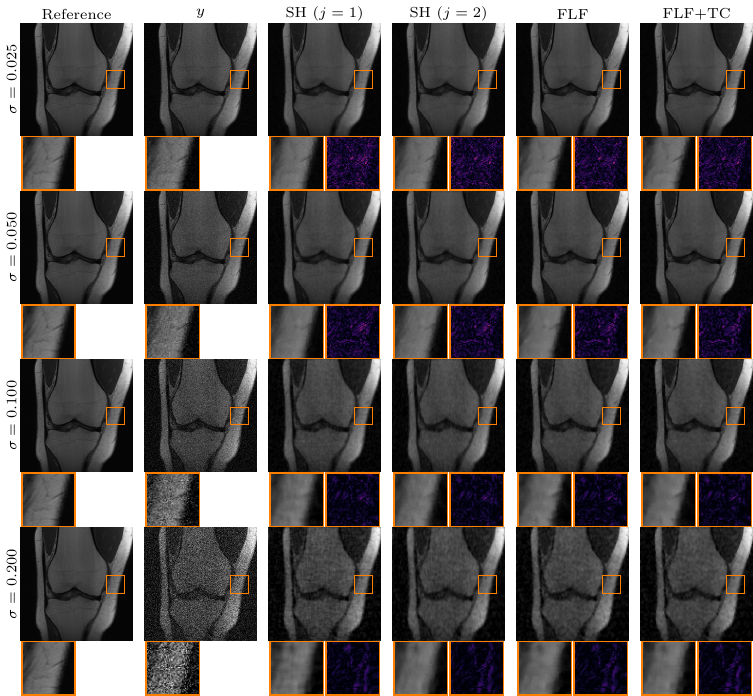}
    \caption{Qualitative denoising results. The noise level increases in each row.
	The inset shows a zoom on the image with corresponding absolute error.
	Our methods: Shearlet (SH), fully-learned filters (FLF) and fully-learned filters and time conditioning (FLF+TC). The number of shearlet scalings is denotes as \( j \).}
	\label{fig:denoising_results_qualitative}
\end{figure*}

The results show that models with learnable convolutional filters consistently outperform those based on shearlet-filters. This also partially explains the performance increase in the~\gls{mri} reconstruction problem. For low and medium noise levels \( \sigma \in \{0.025, 0.050, 0.100\} \), the \gls{psnr} improvement is close to \( 1 \) dB for the model that additionally learns the time conditioning on \gls{corpd}. At a higher noise level \( \sigma=0.200 \) the \gls{psnr} gap between the best shearlet- and fully-learned convolutional model decreases to roughly \( 0.5 \) dB. Overall, these findings demonstrate that learned \glspl*{pogmdm} act as effective denoisers across a range of noise scales, and that the denoising performance increases when further relaxing the theoretical assumptions discussed in \Cref{ssec:GMDiffusion}. We provide qualitative results supporting the reported performance metrics in~\Cref{fig:denoising_results_qualitative}.

\section{Deep J-Sense Details}
\label{sec:appendix_deep_jsense_training}
In this section, we provide additional details on Deep J-Sense~\cite{arvinte2021deepjsense} and the training setup. The method models the image \(\mathbf{x}\) and coil sensitivities \(\mathbf{s}\) directly in k-space, and implements an unrolled alternating reconstruction scheme. Specifically, the updates are given by
\begin{align}
	\mathbf{z} &= \underset{{\mathbf{z}}}{\operatorname{argmin}} \frac12 \norm{\mathbf{y} - \mathbf{A}\mathbf{z}}_2^2 + \lambda_{\mathbf{z}} \norm{\mathbf{F}D_{\theta, \mathbf{s}}(\mathbf{F}^{H}\mathbf{z}) - \mathbf{z}}_2^2, \label{eq:quadratic_s} \\
	\mathbf{s} &= D_{\theta, \mathbf{s}}(\mathbf{F}^{H}\mathbf{z}), \label{eq:prox_s}\\
	\intertext{and}
	\mathbf{v} &= \underset{{\mathbf{v}}}{\operatorname{argmin}} \frac12 \norm{\mathbf{y} - \mathbf{A}\mathbf{v}}_2^2 + \lambda_{\mathbf{v}} \norm{\mathbf{F}D_{\theta, \mathbf{x}}(\mathbf{F}^{H}\mathbf{v}) - \mathbf{v}}_2^2, \label{eq:quadratic_x} \\
	\mathbf{x} &= D_{\theta, \mathbf{x}}(\mathbf{F}^{H}\mathbf{v}), \label{eq:prox_x}
\end{align}
which is reminiscent of a plug-and-play half quadratic splitting scheme in each variable block~\cite{zhang2017pnp_hqs}. The subproblems in~\eqref{eq:quadratic_s} and~\eqref{eq:quadratic_x} are solved using the~\gls{cg} method, while the updates in~\eqref{eq:prox_s} and~\eqref{eq:prox_x} are implemented via~\gls{cnn} denoisers \(D_{\theta,i}\) with \(i \in [\mathbf{x}, \mathbf{s}]\). The method is unrolled for a fixed number of iterations, and the parameters of the denoisers are shared across iterations.
We train the models \(D_{\theta,\mathbf{x}}\) and \(D_{\theta,\mathbf{s}}\) using the same~\gls{corpd} train split and number of slices as described in \Cref{ssec:exp_data}. Furthermore, we adopt the same network architecture as in~\cite{arvinte2021deepjsense}. We use six unrolled iterations and set the number of \gls{cg} iterations to three per block. The parameters are optimized by minimizing an~\gls{ssim} loss using the Adam optimizer~\cite{kingma2015adam} with a learning rate of \(2 \times 10^{-4}\) for 30 epochs. We use the publicly available training code provided by the authors.\footnote{\url{https://github.com/utcsilab/deep-jsense}.}.

In the following, we further examine the out-of-distribution results of~\gls{djs} reported in~\Cref{ssec:parallel_imaging}. To this end, we investigate boxplots of the~\gls{psnr} over all test examples for~\gls{corpdfs}, considering both T1 and T2 contrasts.

\begin{figure*}
	\includegraphics{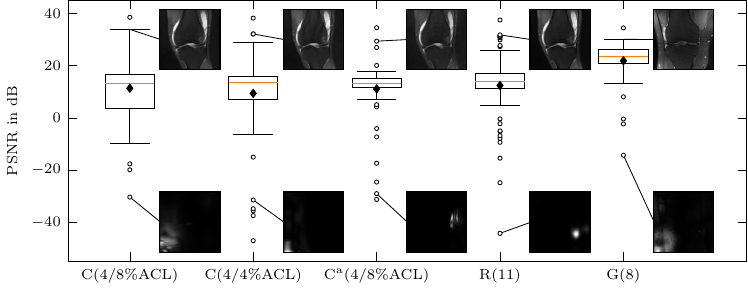}
	\caption{Boxplot of the PSNR values obtained by Deep J-Sense~\cite{arvinte2021deepjsense} on the~\gls{corpdfs} dataset across all k-space trajectories. The median is marked in orange, the mean as \(\blacklozenge\) and the outlier as \(\circ\). We show extreme examples of reconstructions for each k-space trajectory on the right. We denote the k-space trajectories as their abbreviation followed by the acceleration and the number of \glspl{acl} for Cartesian k-space trajectories. The superscript \textsuperscript{a} indicates a rotated phase-encoding direction. Abbreviations: C: Cartesian, R: Radial, G: 2D Gaussian, A: Acceleration, ACL: Auto Calibration Lines.}
	\label{fig:boxplot_djs_corpdfs}
\end{figure*}

\begin{figure*}
	\includegraphics{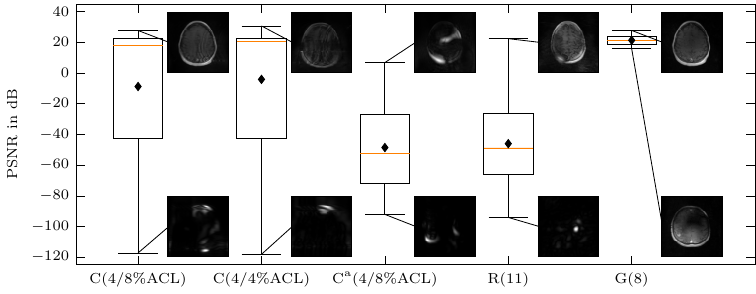}
	\caption{Boxplots of the PSNR values obtained by Deep J-Sense~\cite{arvinte2021deepjsense} on the T1 brain dataset across all k-space trajectories. The median is marked in orange, the mean as \(\blacklozenge\) and the outlier as \(\circ\). We show extreme examples of reconstructions for each k-space trajectory on the right. We denote the k-space trajectories as their abbreviation followed by the acceleration and the number of \glspl{acl} for Cartesian k-space trajectories. The superscript \textsuperscript{a} indicates a rotated phase-encoding direction. Abbreviations: C: Cartesian, R: Radial, G: 2D Gaussian, A: Acceleration, ACL: Auto Calibration Lines.}
	\label{fig:boxplot_djs_t1}
\end{figure*}

\begin{figure*}
	\centering
	\includegraphics{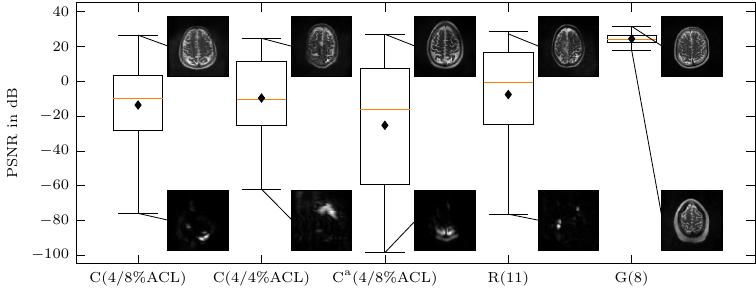}
	\caption{Boxplots of the PSNR values obtained by Deep J-Sense~\cite{arvinte2021deepjsense} on the T2 brain dataset across all k-space trajectories. The median is marked in orange, the mean as \(\blacklozenge\) and the outlier as \(\circ\). We show extreme examples of reconstructions for each k-space trajectory on the right. We denote the k-space trajectories as their abbreviation followed by the acceleration and the number of \glspl{acl} for Cartesian k-space trajectories. The superscript \textsuperscript{a} indicates a rotated phase-encoding direction. Abbreviations: C: Cartesian, R: Radial, G: 2D Gaussian, A: Acceleration, ACL: Auto Calibration Lines.}
	\label{fig:boxplot_djs_t2}
\end{figure*}

From the boxplots in~\Cref{fig:boxplot_djs_corpdfs,fig:boxplot_djs_t1,fig:boxplot_djs_t2}, we again observe that the performance of~\gls{djs} on~\gls{corpdfs} and T1/T2 brain data is highly variable across k-space trajectories. While some reconstructions are plausible, and achieve high \gls{psnr} values, others collapse to artifacts. This variability also explains the extreme~\gls{nmse} values observed in~\Cref{tab:pi_corpdfs,tab:pi_braint1,tab:pi_braint2}, as poorly reconstructed images contribute disproportionately to the overall error. 
For~\gls{corpdfs}, we observe this deterioration in particular when the data contains high noise. We suspect, that this is the case because these examples strongly differ from the data seen during the training of the denoisers. For T1- and T2-weighted brain data, we observe that the method can produce plausible reconstructions, especially for T2-weighted contrast. However, it fails drastically for some examples of the test set which results in high negative~\gls{psnr} values that in turn lower average downstream performance.

In the following, we present additional evaluation results of a Deep J-Sense model trained on a subset of the fastMRI~\cite{knoll2020fastmri} dataset containing both~\gls{corpd} and~\gls{corpdfs} scans. The training set comprises 489~\gls{corpdfs} and 484~\gls{corpd} scans. For each scan, we use the central five slices, resulting in a total of 4,865 slices for training. We can see, that the model works well on~\gls{corpd} data and improves the results on~\gls{corpdfs}. However, the evaluation metrics on~\gls{corpdfs} remain highly variable and show high~\gls{nmse}. One reason for this could be the high noise in the~\gls{corpdfs} training data and the low number of~\gls{cg} iterations.

\begin{table}[!h]
\centering
\caption{Quantitative evaluation of Deep J-Sense (DJS)~\cite{arvinte2021deepjsense} trained on both~\gls{corpd} and~\gls{corpdfs} data. 
The rows alternate between \gls{psnr}, \gls{ssim} and \gls{nmse}. The \gls{nmse} is scaled by $10^2$, and we denote $\times 10^x$ as $ex$. All metrics are shown as mean $\pm$ unit standard deviation. Bold typeface indicates the best method.}
\begin{tabular*}{\textwidth}{@{\extracolsep\fill}ccccccc} 
\toprule
\multirow{3}{*}{T} &
\multirow{3}{*}{A} &
\multirow{3}{*}{\gls{acl}} &
\multicolumn{2}{c}{\gls{corpd}} &
\multicolumn{2}{c}{\gls{corpdfs}} \\
\cmidrule(lr){4-5}
\cmidrule(lr){6-7}
& & & ZF & DJS & ZF & DJS \\
\midrule
\multirow{9}{*}{C}
& \multirow{9}{*}{4} & \multirow{3}{*}{8} 
& $27.19 \pm 1.85$ & $34.49 \pm 2.46$ & $26.32 \pm 2.75$ & $31.34 \pm 5.22$ \\
& & & $0.74 \pm 0.04$ & $0.90 \pm 0.04$ & $0.68 \pm 0.08$ & $0.76 \pm 0.14$ \\
& & & $2.37 \pm 0.72$ & $0.45 \pm 0.24$ & $7.08 \pm 4.07$ & $12.89 \pm 84.29$ \\
\cmidrule(l){3-7}
& & \multirow{3}{*}{8\footnotemark[1]} 
& $31.13 \pm 2.36$ & $34.39 \pm 3.16$ & $26.68 \pm 3.14$ & $16.06 \pm 12.69$ \\
& & & $0.81 \pm 0.05$ & $0.92 \pm 0.04$ & $0.71 \pm 0.08$ & $0.43 \pm 0.31$ \\
& & & $0.97 \pm 0.40$ & $0.51 \pm 0.49$ & $7.03 \pm 4.95$ & $9.5e4 \pm 5.0e5$ \\
\cmidrule(l){3-7}
& & \multirow{3}{*}{4} 
& $24.14 \pm 1.66$ & $33.92 \pm 2.46$ & $25.19 \pm 2.33$ & $31.26 \pm 3.81$ \\
& & & $0.69 \pm 0.04$ & $0.90 \pm 0.04$ & $0.66 \pm 0.08$ & $0.77 \pm 0.12$ \\
& & & $4.96 \pm 1.52$ & $0.52 \pm 0.34$ & $9.05 \pm 4.23$ & $10.72 \pm 65.01$ \\
\midrule
\multirow{3}{*}{R} 
& \multirow{3}{*}{11} & \multirow{3}{*}{-} 
& $28.76 \pm 2.15$ & $33.15 \pm 2.29$ & $25.25 \pm 3.25$ & $28.13 \pm 5.90$ \\
& & & $0.75 \pm 0.06$ & $0.86 \pm 0.04$ & $0.62 \pm 0.10$ & $0.70 \pm 0.18$ \\
& & & $1.67 \pm 0.58$ & $0.61 \pm 0.29$ & $10.94 \pm 8.48$ & $6.5e2 \pm 4.1e3$ \\
\midrule
\multirow{3}{*}{G} 
& \multirow{3}{*}{8} & \multirow{3}{*}{-} 
& $32.15 \pm 2.32$ & $27.00 \pm 3.76$ & $26.86 \pm 3.84$ & $25.78 \pm 5.53$ \\
& & & $0.84 \pm 0.05$ & $0.82 \pm 0.06$ & $0.70 \pm 0.10$ & $0.70 \pm 0.12$ \\
& & & $0.78 \pm 0.35$ & $4.22 \pm 4.47$ & $7.98 \pm 6.89$ & $8.98 \pm 11.11$ \\
\midrule
\multicolumn{3}{c}{\multirow{2}{*}{\footnotesize \makecell{Number of\\Parameters}}} & \multirow{2}{*}{-} & \multirow{2}{*}{447490} & \multirow{2}{*}{-} & \multirow{2}{*}{447490} \\
& & & & & & \\
\bottomrule
\end{tabular*}
\footnotetext{\textsuperscript{a} Rotated phase-encoding direction.}
\footnotetext{T: k-space Trajectory, C: Cartesian, R: Radial, G: 2D Gaussian, A: Acceleration, ACL: Auto Calibration Lines}
\label{tab:deep_j_sense_extended_results}
\end{table}

\end{appendices}
\newpage
\bibliography{sn-bibliography}

\end{document}